\documentclass[conference]{IEEEtran}

\usepackage{mathrsfs}
\usepackage{multirow}
\usepackage{graphics} 
\usepackage{epsfig} 
\usepackage{mathptmx} 
\usepackage{amsmath} 
\usepackage{amssymb}  
\usepackage{subfigure}
\usepackage{array}
\usepackage{graphicx}
\usepackage{leftidx}
\usepackage{textcomp}
\usepackage{bm}
\usepackage{xcolor}
\usepackage{soul}
\usepackage{diagbox}

\usepackage{times}
\usepackage[numbers]{natbib}
\usepackage{multicol}
\usepackage[bookmarks=true]{hyperref}

\begin{document}

\title{General Robot Dynamics Learning and Gen2Real}


\author{\authorblockN{Dengpeng Xing$^{1}$, Jiale Li$^{1}$, Yiming Yang$^{1}$, and Bo Xu}
\authorblockA{Institute of Automation, Chinese Academy of Sciences, Beijing, China\\
Email: dengpeng.xing@ia.ac.cn, lijiale2019@ia.ac.cn, yangyiming2019@ia.ac.cn, xubo@ia.ac.cn}
}


%

\maketitle

\begin{abstract}
Acquiring dynamics is an essential topic in robot learning, but up-to-date methods, such as dynamics randomization, need to restart to check nominal parameters, generate simulation data, and train networks whenever they face different robots. To improve it, we novelly investigate general robot dynamics, its inverse models, and Gen2Real, which means transferring to reality. Our motivations are to build a model that learns the intrinsic dynamics of various robots and lower the threshold of dynamics learning by enabling an amateur to obtain robot models without being trapped in details. This paper achieves the ``generality'' by randomizing dynamics parameters, topology configurations, and model dimensions, which in sequence cover the property, the connection, and the number of robot links. A structure modified from GPT is applied to access the pre-training model of general dynamics. We also study various inverse models of dynamics to facilitate different applications. We step further to investigate a new concept, ``Gen2Real'', to transfer simulated, general models to physical, specific robots. Simulation and experiment results demonstrate the validity of the proposed models and method.\footnote{ These authors contribute equally.}
\end{abstract}

\IEEEpeerreviewmaketitle

\section{Introduction}

Learning renders robots capable of performing well a variety of tasks in a diversity of environments and has recently attracted worldwide attention \cite{Kaelbling915}. Many types of learning topics are investigated in robotics, such as imitation learning and reinforcement learning, and a good many up-to-date robot mechanisms are involved, such as soft robots \cite{Shah} and legged robots \cite{Hwangbo}. Although their intelligence has been tremendously improved by learning different skills (such as grasping \cite{Ficuciello}, locomotion \cite{Won}, manipulating \cite{Fazeli}), robots cannot step into our daily life soon. One critical difficulty lies in data. Although several documents attempt to collect data directly from physical robots \cite{Levine18}, it is quite costly to acquire sufficient data from the real world, with reasons including enormous variability in the environments, tear and wear, and so forth. Training models or policies in simulations and transferring to the real world is one feasible solution since a simulator theoretically provides enough data. However, it faces the reality gap, which depicts the difference between the simulator and the real world, and many researchers endeavor to close this Sim2Real gap \cite{hofer}. Domain randomization trains cross many environments where parameters and properties are randomized, expecting the physical system as one sample of training variations. This technique is simple yet effective and especially suitable for deep networks. It is applied widely to many robotic tasks and achieves good performance.

For robot learning, dynamics is essential to be concerned, especially for movement planning and policy optimization, relating forces acting on a robot mechanism with accelerations they produce. Model-Based reinforcement learning algorithms, for instance, need to capture dynamics changes, which may fail to provide accurate transition states \cite{moerland2021modelbased}. Many documents study the learning of robot dynamics. System identification is an early attempt by tuning parameters to match robot behaviors \cite{Yu-RSS-17}, and its shortcoming is obvious: time-consuming and error-prone. Dynamics randomization augments the training set with a wide range of parameters for dynamics learning \cite{Peng}. This method has shown good performance in transferring to reality but still needs to generate simulation data and apply the Sim2Real technique whenever it meets new requirements with different robot configurations. This paper considers the following question to spare this tediousness of iterative sampling and modeling: Is it possible to concern general robot dynamics (GRD), with which practitioners can simplify the process of dynamics learning? GRD should cover massive robot instances and have the capability of transferring to a variety of physical mechanisms. This new concept has substantial merits. It implies the essence of the robot dynamics, provides a generalization environment for learning in simulation, and lowers the corresponding threshold, enabling a beginner to obtain robot models without attention to details.

Generative pre-training (GPT) is a transformer-based model trained on a massive dataset, exhibiting significant ability in the area of natural language processing \cite{Brown}. It is similar to decoder-only transformers, with the difference lying in the immense model scale and training data. As an unsupervised learning fashion, this method also applies to predict pixels without structure knowledge \cite{Chen}, label data for the graph neural networks \cite{Hu}, generate synthetic biological signals \cite{Bird}, and so forth. Its success may lie in that instances of downstream tasks appear in the succeeding inputs, which facilitates prediction. We examine that, for any robot model, dynamics property exists in continuous trajectories, where inputs are a series of coherent variables and satisfy GPT's requirements. That makes it possible to use a GPT-similar structure to learn GRD.

This paper studies the learning of GRD, endeavoring to pre-train a model on a variety of serial robots. So far as we know, this is the first concern of learning GRD. We investigate the \emph{generality} from three aspects: Dynamics parameter, topology configuration, and model dimension. The model dimension determines the number of robot variables; the topology configuration determines the connection of robot joints; the dynamics parameter determines the property of robot links. We generate datasets by randomizing the above three aspects and execution process within constraints and apply a structure modified from GPT to access the general dynamics.

We also extend the ``generality'' idea to the inverse of robot dynamics. Based on the correlation with GRD, we study the general inverse dynamics independently and the left and the right inverse of GRD with specific purposes, which consider the errors of GRD.

This paper continues to investigate a new concept, \emph{Gen2Real}, to transfer general models directly to reality. For a clear view, given a physical robot, this process can be divided into Gen2Spe and Spe2Real. The first trains a general model to fit a specific robot in simulation, and the second transfers the simulated model to reality with experiment data.

We hope the general models we provide can facilitate policy learning in the simulation since they include enormous robot dynamics. We also hope, with these models, practitioners can spare the tedious process, in transferring to reality, of checking nominal parameters, generating lots of simulation data, and training networks. In summary, these general models can lower the threshold, attracting beginners in robot learning.

\section{Related Work}

\subsection{Robot dynamics learning}
Robot dynamics is crucial to simulating behaviors in policy learning, and, for example, model-based reinforcement learning needs dynamics to provide state transition whenever it attempts a policy. The dynamics cannot yet be accurate due to ubiquitous noises and disturbances. Many learning methods are presented to address it, such as reusing existed experience \cite{Christopher}, employing an episodic method \cite{Folkestad}, exploiting Hopf bifurcations \cite{Khadivar}, recruiting recurrent spiking neuron networks \cite{Gilra}, and so forth. The inverse dynamics also attracts attention in designing appropriate motion, and an example is the design of model-based controllers, which cancel out non-linearities and track with zero-error. It also bridges the torques of the simulated model and the states of the physical robot \cite{Desai}. Some identification-based methods are applied to learn inverse dynamics of robot manipulators, such as a cascaded method imitating the recursive Newton-Euler formulation \cite{Sahand}, combining online and offline neural networks \cite{Panda}, learning a non-minimum phase system \cite{Zhou}. These models are good at learning one specific robot and not suitable for directly transferring to another physical platform.

\subsection{Domain randomization}

Domain randomization varies parameters to define a simulation environment and models differences between the source and the target domains. It employs a wide range of simulated parameters and intends to improve neural networks to generalize well to real-world tasks. Many vision-based policy tasks apply this method and train networks on simulation data. The randomization may include scene appearance and robot kinematics, and visual control policies are learned and transferred to the non-randomized real-world images. Recent techniques utilize this randomization to deal with partial occlusions \cite{Tobin}, adapt distribution with a few real-world data \cite{Chebotar}, and use 3D CAD images as the source inputs \cite{Sadeghi}.

When solving dynamics-related tasks, dynamics randomization is a prevalent method to generalize skills to reality \cite{Valassakis}. The physical dynamics features may include mass and sizes of robot bodies, joint friction, observation noise, and other properties. It applies to develop policies by randomizing simulated dynamics \cite{Peng}, learn predictive dynamics in inference \cite{Alvaro}, and design a universal policy based on training over a wide array of dynamics models \cite{Wenhao}. Another successful application is dexterous manipulation \cite{Andrychowicz}, in which parameters with high uncertainty are frequently randomized. One view to the success of domain randomization is that it is based on bilevel optimization.

Its main challenge lies in selecting the proper parameter set, which is sensitive to the manually-specified distribution \cite{Liang-RSS-20}. One approach to solving the problem is active domain randomization \cite{Mehta}, which learns a sampling strategy of parameters, searching for the most informative variations within randomization ranges. Another is automatic domain randomization \cite{Akkaya}, which attempts to generate randomized environments automatically with changeable distribution ranges.

In the existing documents relating to dynamics tasks, most are focused on the randomization of dynamics parameters considering the target robot, which works well for this type of physical mechanism, and randomization for other influential factors to robot dynamics, such as topology configurations and model dimensions, is not mentioned.

\subsection{Generative pre-training}
The technique of large-scale pre-training uses architectures based on transformers and recently achieves impressive success. GPT-2 \cite{Radford}, as an example, can improve natural language understanding on a diverse corpus of unlabeled text after training on an enormous dataset. GPT-3 \cite{Brown} is proposed recently to use 175 billion parameters and 570GB of training data and achieves strong performance on many datasets without gradient updates. These models demonstrate the power of GPT and have been applied to many other disciplines. It is used to predict pixels knowing nothing about the input structure \cite{Chen}, learn speech representations \cite{Chung}, classify tumors in MR images \cite{Ghassemi}, and so forth. It is also employed to overcome defects of other methods, e.g., to reduce the labeling effort of graph neural networks \cite{Hu}.

\section{General Dynamics Model}
We first analyze the influential factors to robot dynamics and then describe the method to learn GRD, with an attempt to cover the dynamics of enormous serial robots. We also discuss the differences from dynamics randomization.
\subsection{Robot dynamics}
Consider robot dynamics
\begin{equation}\label{equ:dynamics}
\textbf{\emph{M}}\left(\textbf{\emph{q}}\right)\ddot{\textbf{\emph{q}}} + \textbf{\emph{C}}\left( \textbf{\emph{q}}, \dot{\textbf{\emph{q}}} \right)\dot{\textbf{\emph{q}}} + \textbf{\emph{G}}\left( \textbf{\emph{q}} \right) = \tau + \textbf{\emph{f}}
\end{equation}
where $\textbf{\emph{q}}$, $\dot{\textbf{\emph{q}}}$, and $\ddot{\textbf{\emph{q}}}$ are the vectors of joint position, velocity, and acceleration, respectively, $\tau$ is the input joint torque, $\textbf{\emph{f}}$ is the joint torque due to friction and disturbances, $\textbf{\emph{M}}$ is the inertia matrix, $\textbf{\emph{C}}$ is the centrifugal and Coriolis item, and $\textbf{\emph{G}}$ is the gravity. Label $\textbf{\emph{s}} = \left[\textbf{\emph{q}}^{\mathrm{T}}, \dot{\textbf{\emph{q}}}^{\mathrm{T}}\right]^{\mathrm{T}}$ to represent the robot state. The dynamics learning is to obtain the mapping from the current state $\textbf{\emph{s}}_t$ and the torque $\tau_t$ to the next state $\textbf{\emph{s}}_{t+1}$. Viewing the above equation, we conclude the following three facts:
\begin{itemize}
\item [1)] The value of each element in those matrices and vectors varies as robot parameters change.
\item [2)] The expression of each matrix is also different as the robot configuration changes.
\item [3)] The dimension of each matrix and vector alters as the robot link number changes.
\end{itemize}
It appears that the configuration and the dimension have more influence since they, other than parameters, change the expression of the dynamics equation.

System identification tunes parameters to match robot behaviors, which corresponds to approximate the exact dynamics equation. It appears to be time-consuming and error-prone. Dynamics randomization augments the training set with a wide range of parameters for dynamics network learning, which considers the variation of matrix values in Eq. \ref{equ:dynamics}. This method needs to restart the whole training process whenever it meets new requirements with different robot configurations. Another possible approach steps further concerning the variation of matrix value, expression, and dimension. It views all the above three factors, and we call it GRD since this model covers cross various dynamics parameters, topology configurations, and model dimensions. This general model builds a general environment for learning in simulation, avoids iterative sampling and modeling, and simplifies dynamics learning.

\subsection{Learning of general dynamics}
\begin{figure}[t]
 \centerline{\psfig{file=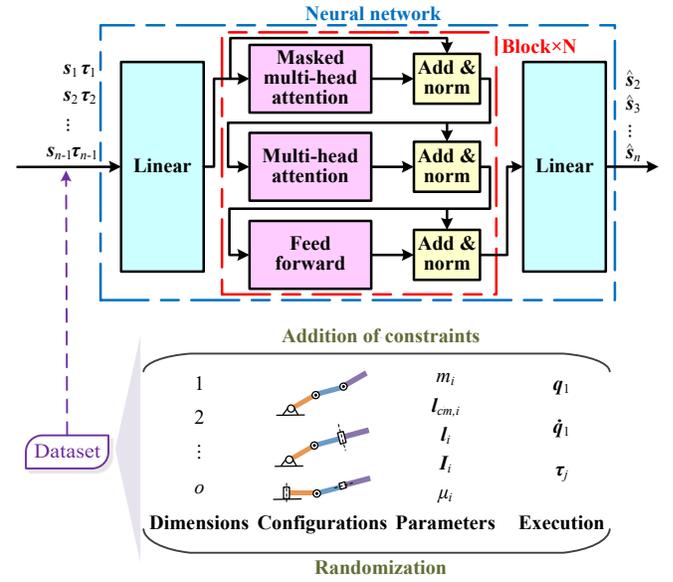,scale=0.95}}
 \caption{The network structure and dataset for GRD learning. The data contains plentiful trajectories, each acquired by randomizing dynamics parameters, topology configurations, model dimensions, initial states, and driving torques. Constraints are added in states and torques to produce meaningful trajectories. The network is modified from GPT to adapt to robot dynamics learning. The inputs are joint states and torques of a trajectory in sequence, and the outputs are the prediction of the succeeding states.} \label{fig:gpt}\vspace{-0.5cm}
\end{figure}

With the above analysis, we present the learning process, as shown in Fig. \ref{fig:gpt}, where a network modified from GPT is employed to learn general dynamics with a dataset. The dataset includes enormous robot models, acquired by randomizing dynamics parameters $P_p$, topology configurations $P_c$, and model dimensions $P_d$. Label $P=\{P_p,P_c,P_d\}$ the robot model.

The model dimensions illustrate how many links a robot has and significantly affect the scale of the dataset. We express it in the form of $P_d\in\{ 1,2,\cdots,o \}$, where $o\in \mathbb{N}^\star$. The results show that the dynamics we learn is downward compatible, which means the model trained for three links can be used directly for 2-link robots. Therefore, it is preferable in the dataset to have more trajectories of robots with more dimensions. 

The topology configurations primarily address the connection type of robot links, namely the setting of each joint. The relative movements between two adjacent links can be rotational, linear, spherical, helical, and so forth. We focus on rotational joints, the elements of articulated robots, for detailed discussion, and our method also applies to other connection types. A revolute pair performs distinct behaviors rotating around a different axis, and we can randomly select the rotational direction, with each joint spinning around an arbitrary axis. However, this arbitrariness does not comply with commonly-used industrial robots, and one can further limit this randomization by setting $P_c=\{a_i|a_i\in\{\pm \textbf{\emph{x}}, \pm \textbf{\emph{y}}, \pm \textbf{\emph{z}}\}\}$. This reduced range means that there is a state at which the joint spins along an axis of the Cartesian coordinates.


The dynamics parameters describe the property of each link, which include mass $m_i$, the center of mass $\textbf{\emph{l}}_{cm,i}$, length $\textbf{\emph{l}}_i$, the moment of inertia $\textbf{\emph{I}}_i$, and friction coefficient $\mu_i$ of link $i$, expressed as $P_p = \{m_i, \textbf{\emph{l}}_{cm,i}, \textbf{\emph{l}}_i, \textbf{\emph{I}}_i, \mu_i\}$. This paper only considers serial robots, and the proposed method can also be available to soft robots, parallel robots, and robots with closed-loop chains. We set dynamics parameters in a manner to cover those commonly used robots. Meanwhile, we can add constraints to parameters to build models closer to real robots. These constraints limit the range of some parameters by relating them together, e.g., the center of mass is not at the end of the link. In other words, the addition of constraints is to exclude the improbable parameters, enhancing the dataset effectiveness. It is worthy to note that a substantial range of parameters is necessary if one aims at a GRD model.


Constraints of joint states and torques are also valuable in generating meaningful movements, which is helpful to prohibit undesired behaviors, such as out-of-reach acceleration and high-speed rotation. To obtain a simulation trajectory, we sequentially randomize a model dimension, a topology configuration, and dynamics parameters, randomly pick an initial state $\textbf{\emph{s}}_1$ under this robot model $P$, and continuously move motors with randomized torques (motor bubbling). Iteratively repeating the above steps leads to a dataset, in which a trajectory takes the form of $\left\{ \textbf{\emph{s}}_1, \tau_1, \cdots, \textbf{\emph{s}}_n, \tau_n\right\}$, where $n$ is the state number.

A substantial range of dynamics parameters and the addition of randomization of topology configurations and model dimensions naturally render a large-scale network structure and a massive dataset for learning. We examine the robot motion sequence in the dataset and find the trajectory characteristics are the effects of the current torque appear in the next state. It satisfies GPT's requirements, and then we use a structure, shown in Fig. \ref{fig:gpt}, modified from GPT, whose core idea is to approximate models via predicting data. The network input is a complete trajectory. More specifically, we can view it as a variable-size matrix, where each row corresponds to a vector $[\textbf{\emph{s}}_j^{\mathrm{T}},\tau_j^{\mathrm{T}}]^{\mathrm{T}}$ with $j\leq o$, and the column consists of each time step of a trajectory in sequence. It is worthy to note that the network is downward compatible: It can accept inputs with model dimensions less than $o$. The network output is also a matrix, where each row is the predicted state of the next time step, and the column corresponds to the time sequence.

We use the root mean square error (RMSE) to represent the deviation of outputs to the succeeding inputs and the self-supervised learning technique for training. The network structure is different from GPT in three aspects. We replace the embedding layer with a fully-connected one for encoding since inputs are vectors, remove the softmax in the output layer to accustom to our tasks, and erase the positional encoding in the front layers (which, we believe, is not necessary for processing motion sequence data). The remaining parts are the same as GPT. $N$ blocks are sequentially connected to improve data processing, and each block is composed of a masked multi-head attention layer, a multi-head attention layer, two feedforward layers, and three ``add \& norm'' modules.  A fully-connected layer follows after the last block. Before processing a state, the multi-head attention improves the relationship understanding of other associated states, with joint attendance to different representation information at various positions \cite{Vaswani}. The ``masking'' aims to mask latter states so that the network predicts only according to current states.

\subsection{Difference from dynamics randomization}
Although both GRD and dynamics randomization appears similar in applying the randomization technique, they possess substantial differences:
\begin{itemize}
\item For research motivation: We intend to study a general dynamics model, aiming at covering various common instances of serial robots, whereas dynamics randomization focuses on a specific robot. By this means, we can directly apply the same general dynamics model to a variety of robots, but dynamics randomization needs to restart generating data and training networks according to newly designated platforms.
\item From the perspective of configuration, we add randomization of joint pins, i.e., to cross over various topology configurations, which props up the idea of generality. Dynamics randomization aims at a specific robot acquired from the target and obviously can not transfer to robots with different configurations.
\item In terms of the parameter: Dynamics randomization mainly focuses on a small range around nominal parameters, e.g. $\left[ 0.25, 4\right]\times$ default mass in \cite{Peng}, while our model faces a substantial parameter range to cover various robot sets. The general dynamics model comparably has more massive parameter ranges.
\item On the aspect of dimension, our model is compatible with robots of various link numbers. It can apply to both planar 2-link robots and cubic 6-link robots. Dynamics randomization does not address this issue and can only apply to the learned dimension.
\item Viewing the network size, we can easily distinguish the two types of networks. Dynamics randomization usually adopts a small network, and an example is the four hidden layers, each containing 128 neurons \cite{Peng}, while our model employs $10^8$-level sized parameters.
\end{itemize}

\section{Inverse Models of General Dynamics}
\begin{figure}[t]
\subfigure[]{
\label{fig:inverse:a}
\begin{minipage}[b]{0.445\textwidth}
\centering
\includegraphics[scale=0.85]{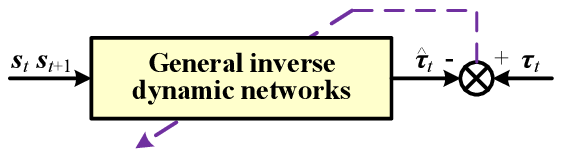}
\end{minipage}}\\
\subfigure[]{
\label{fig:inverse:b}
\begin{minipage}[b]{0.22\textwidth}
\centering
\includegraphics[scale=0.85]{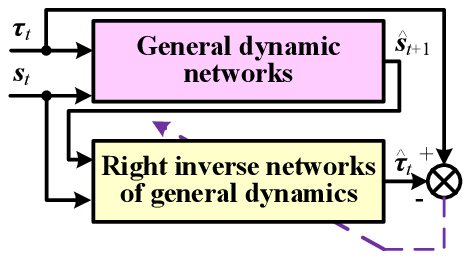}
\end{minipage}}
\subfigure[]{
\label{fig:inverse:c}
\begin{minipage}[b]{0.22\textwidth}
\centering
\includegraphics[scale=0.85]{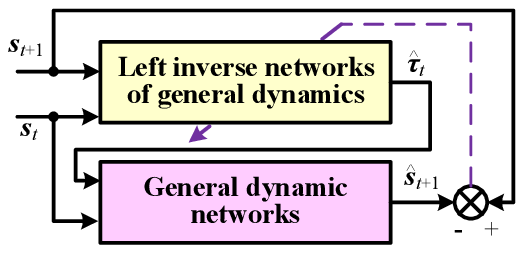}
\end{minipage}}
\caption{Inverse dynamics related models. (a) The general inverse dynamics model, which is trained independently from the dynamics. (b) The right inverse networks and (c) the left inverse networks of GRD consider the dynamics error in its learning. The first network has distinct physical meanings, and the latter ones form autoencoders with GRD.}
\label{fig:inverse}\vspace{-0.5cm}
\end{figure}

The state transition of various robots is acquired after learning GRD, and similarly, we can investigate its inverse models. Inverse dynamics accepts the current state, $\textbf{\emph{s}}_t$, and the desired next state, $\textbf{\emph{s}}_{t+1}$, and outputs the torque, $\tau_t$, to achieve that state transition. There are two types of models relating to inverse dynamics, as shown in Fig. \ref{fig:inverse}, where the upper one is to learn general inverse dynamics and the lower two are to learn inverse models of GRD. For the learning of either model, the same dataset described in the previous section is used.

One can refer to the same network structure as is applied to learn GRD to acquire general inverse dynamics, as shown in Fig. \ref{fig:inverse:a}, which also covers massive robot instances. Unlike the learning process of GRD, the outputs of the inverse dynamics do not appear in the succeeding inputs, and thus supervised learning is applied to train the network. This model has distinct physical meanings and can be used in a pre-trained environment, e.g., trajectory planning under torque constraints and trajectory tracking.

Since general models still have non-negligible learning errors even after sufficient training, the independently-trained inverse dynamics has no (apparently physical) relationship with GRD. It will produce larger (about twice of) model errors if putting them together. To solve this problem, we view it from another aspect, studying the inverse of general dynamics considering both models' errors. The lower half in Fig. \ref{fig:inverse} presents two inverse models, which are diverse as GRD produces apparent errors, and the connection of general dynamics and its inverse forms an autoencoder, limiting the gain of the two networks around one. In one viewpoint, the inverse models intend to reduce the errors produced by GRD from different directions. The right inverse in Fig. \ref{fig:inverse:b} takes the current state and the outcomes of GRD as input and endeavors to generate torques that equal the inputs to GRD. The combination of dynamics and its right inverse can apply to circumstances in which torques are supposed to be maintained or tracked. Given the current and succeeding states, the left inverse in Fig. \ref{fig:inverse:c} aims to result in torques with which GRD can output in a small range around the succeeding state. This combination of left inverse and GRD is valuable in planning and tracking trajectories. The self-supervised learning technique is applied to train the two inverse models while unchanging the dynamics model.

The differences of three inverse models in Fig. \ref{fig:inverse} are: Given the current state $\textbf{\emph{s}}_t$, the inverse dynamics maps the succeeding state $\textbf{\emph{s}}_{t+1}$ to the torque $\tau_t$; the right inverse model endeavors to relate the output of GRD $\hat{\textbf{\emph{s}}}_{t+1}$ with the torque $\tau_t$; the left inverse maps the succeeding state to the torque $\hat{\tau}_t$ that enables GRD to generate this succeeding state. With errors from GRD, these inverse models perform differently but become similar when GRD's errors approach acceptable values.

\section{Gen2Real}
GRD and its inverse models provide a simulation environment covering massive robot models with variant parameters, configurations, and dimensions, in which one can test and optimize his/her policies. After that, it is desirable to transfer models and skills to physical robots, and here we introduce a new concept, Gen2Real, which bridges between general models and reality.

\begin{figure}[t]
 \centerline{\psfig{file=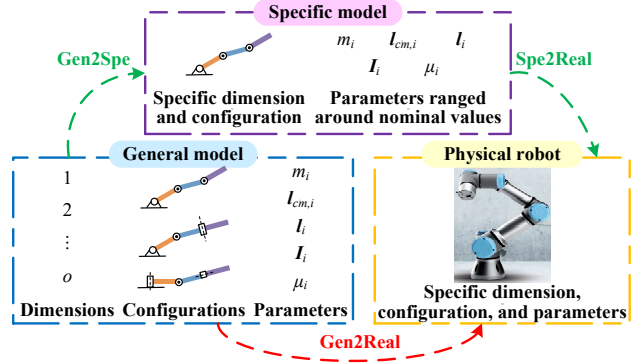,scale=0.9}}
 \caption{The two processes of transferring to reality. One is Gen2Real, which transfers GRD to reality with experiment data. This process is simple, and practitioners can directly use it without trapped into details. Another is the combination of Gen2Spe and Spe2Real. Gen2Spe transfers GRD to a specific model with simulation data. This new model has the same dimension and configuration as the target robot, and the parameters are randomized in a range around the nominal values. Spe2Real transfers the specific model to reality with experiment data.} \label{fig:gen2real}\vspace{-0.5cm}
\end{figure}

There are typically two ways, as shown in Fig. \ref{fig:gen2real}, for model transfer. One is Gen2Real, transferring simply from general dynamics to reality (a determined robot target) with experimental data. We can view it as a transfer process from a simulated model with a randomized setting (including parameters, configurations, and dimensions) to a physical, specific robot. It is convenient: Anyone can directly apply our pre-trained general models to different robots with experiment data generated by motor bubbling and obtain dynamics networks with sufficient fitting precision, which spares the dull modeling and training in simulations (as what researchers do in Sim2Real).

Another transferring method is a combination of Gen2Spe and Spe2Real processes, as the dashed green arrows shown in Fig. \ref{fig:gen2real}. Gen2Spe aims to adapt a general model toward a specific one that is closer to the target robot. To do it, we need to generate a simulation dataset by applying the same configuration and dimension with the target robot and randomizing dynamics parameters in a range around nominal values, and then train the network to fit the simulated robot model. Spe2Real transfers the specific model to reality with experiment data, which is similar to the process commonly used in dynamics randomization, with the difference lying in the network structure and size. The specific model is like a transfer station in a bridge between general models and reality, whose purpose is theoretically to increase the transferring performance at the expense of the addition of simulation data.

Both two methods have merits. Gen2Real is direct, and practitioners can merely apply it without worrying about details, such as the robot's configuration and parameters. Gen2Spe and Spe2real take a roundabout route, requiring both training in simulation and the robot's nominal parameters, and perform well if the physical robot is a sample of the specific model.

Due to the massive network size, adjusting all weights during transferring will not receive good results, given a small number of experiment data, and correspondingly we only tune a few layers, which performs well. Pruning as transferring will be considered in later research.

\section{Experiments and Results}
\subsection{The pre-trained general models}

We apply an iterative process of setting a model and running trajectories to generate a dataset for learning in simulation. For the model set, we sequentially randomize a model dimension with $o\leq 6$, an axis direction for each revolute joint, and dynamics parameters for each link. We add several constraints to limit the dataset size in its generation, and the proposed method can apply to other instances out of the limitations. The setting of no-more-than-six dimensions is to consider the minimum requirement of freely moving in three-dimensional space. The consideration of revolute joints is to focus on the connection of articulated robots. Tab. \ref{tab:I} shows the parameter range, which considers both the commonly used robot models and the data effectiveness. We further strengthen the parameter constraints by relating some parameters: The relationship between link length and the center of mass avoids lopsided links, and the moment of inertia relates with mass and link length to strict the material density in a reasonable range. It shows that the parameter range for GRD is much broader than used in dynamics randomization. After acquiring this robot model, we sequentially randomize an initial state, $\textbf{\emph{s}}_1$, and joint torques for 50 time-steps in one trajectory. During execution, based on Eq. \ref{equ:dynamics}, we use fourth-order Runge-Kutta to integrate motion in a time interval $\Delta t = 0.1s$. This paper uniformly samples in randomization ranges, and other sampling methods, such as logarithmical sampling, can also be used. In total, the dataset has 20 million trajectories, each having 50 state points. We use $90\%$ for training and $10\%$ for testing.

\begin{table}[t]
\newcommand{\tabincell}[2]{\begin{tabular}{@{}#1@{}}#2\end {tabular}}
\caption{The range of dynamics parameters, initial states, and driving torques.} \label{tab:I}
\begin{center}
\renewcommand\arraystretch{1.5}
\begin{tabular}{p{2.5cm}<{\centering} p{3.5cm}<{\centering} }
\hline\hline
Parameter & Range  \\
\hline
Mass $m_i$ & $\left[ 0.1,10\right]kg$ \\

Friction coefficient $\mu_i$ & $[0.5,2.5]$ \\

Center of mass $l_{cm,i}$ & $[-0.1,0.5]m$ \\

Link length $l_i$ & $l_{cm,i}\times [\frac{10}{7}, \frac{10}{3}]m$ \\

Moment of Inertia $I_i$ & $\frac{1}{12}m_i\times l_i^2\times[0.3,3.0]kg\cdot m^2$ \\

Initial position $q_{i,1}$ & $[-\pi,\pi] radians$ \\

Initial velocity $\dot{q}_i$ & $\left[-1,1\right] rad/s$ \\

Joint Torque $\tau_i$ & $[-30,30]N\cdot m$\\
\hline\hline
\end{tabular}
\end{center}
\end{table}

Tab. \ref{tab:II} shows the GRD learning size and corresponding performance. In this table, $n_{params}$ is the total parameters; $n_{blocks}$ is the block number; $d_{model}$ is the data dimension of each block and also indicates the neuron number of linear layers, where the first layer includes $d_{model}$ neurons and the second incorporates $4\times d_{model}$ neurons; and $n_{heads}$ is the head number of the ``multi-head'', which reshapes a long vector into matrices. We display two general dynamics models, i.e., 2-link and 3-link robots, to show the learning performance. For the learning structure of 3-link dynamics, as shown in Tab. \ref{tab:II}, we provide 200 million parameters in total, consisting of 64 blocks with eight heads and 512 data dimensions in each block. Its RMSE is 0.1557, which contains the errors of joint positions and velocities. We also take the 3-link model as an example to test the effect of structure scale. The medium structure contains half parameters, and the small one has quarter parameters. We change their block number without revising each block. The results show that the increase of structure scale is helpful to the performance of general dynamics.

\begin{figure}[t]
\subfigure[]{
\label{fig:dynamics_result:a}
\begin{minipage}[b]{0.23\textwidth}
\centering
\includegraphics[scale=0.33]{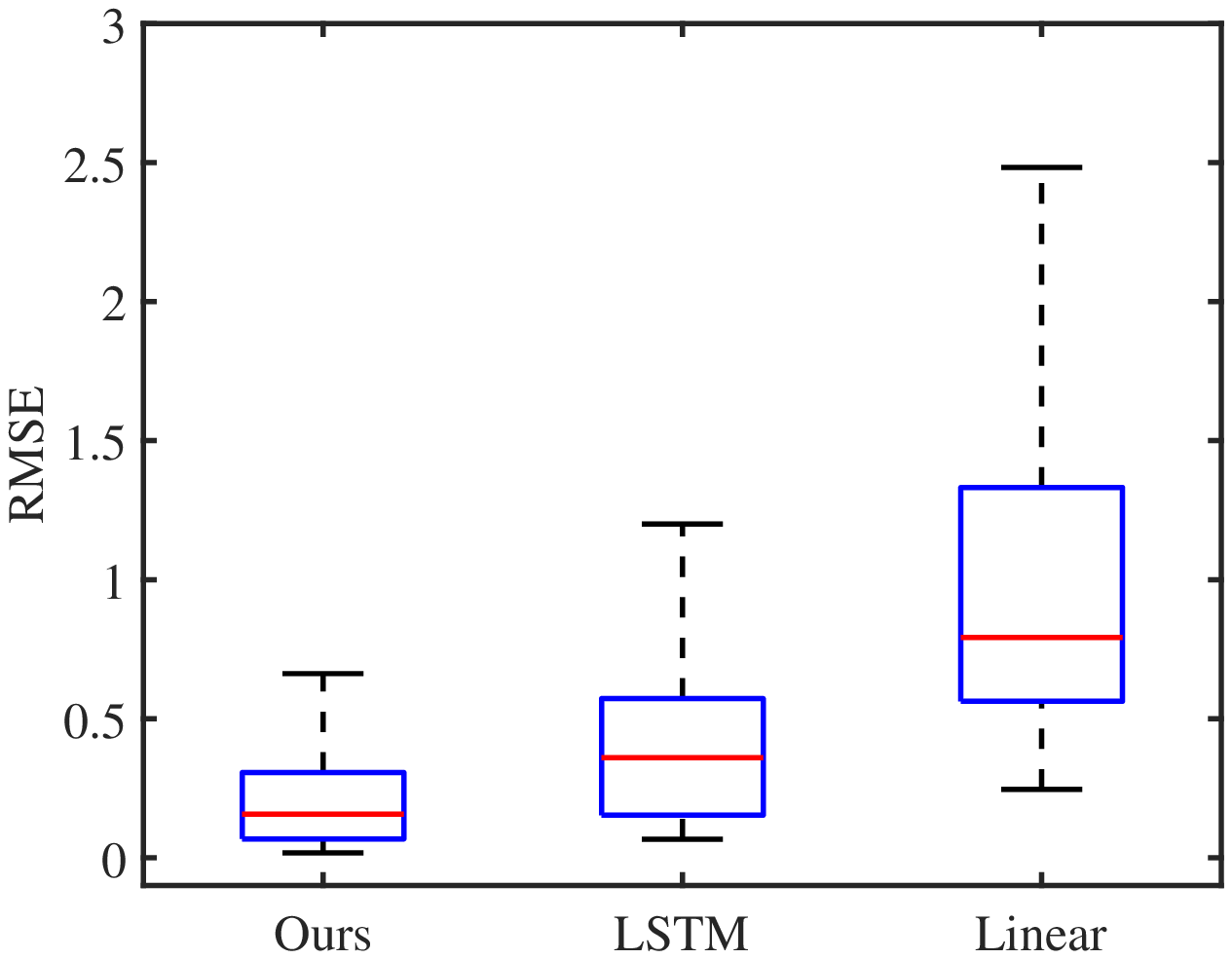}
\end{minipage}}
\subfigure[]{
\label{fig:dynamics_result:b}
\begin{minipage}[b]{0.23\textwidth}
\centering
\includegraphics[scale=0.33]{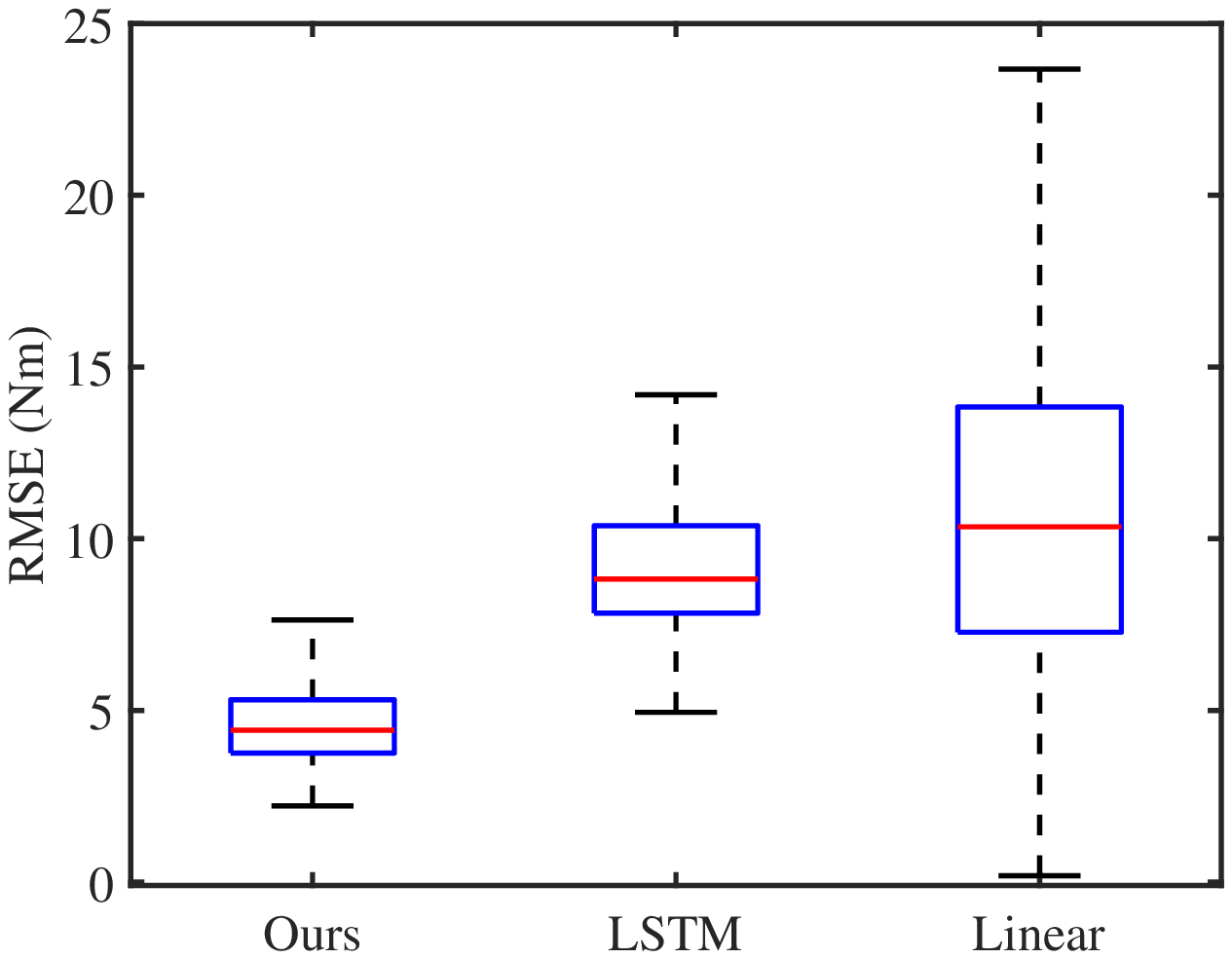}
\end{minipage}}
\caption{The error comparison of different methods in tracking tasks with various robot a) dynamics and b) inverse dynamics.}
\label{fig:dynamics_result}
\end{figure}

We compare the performance of approximating various robot models with other approaches. To do it, we generate five thousand trajectories using both 2-link and 3-link models and test the performance in tracking tasks. We pick two comparative methods: An LSTM with $4\times 128$ neurons employed in \cite{Peng} and a 5-layer linear network with about 7200 neurons. Fig. \ref{fig:dynamics_result:a} shows the results. The linear network has the largest errors, with a mean of 1 and a standard deviation of 0.58, and the LSTM is much better, with a half mean and a half standard deviation. Our model performs the best on these trajectories, with a mean of 0.17 and a standard deviation of 0.16. These results demonstrate the advantages of our general dynamics model.

\begin{table}[t]
\newcommand{\tabincell}[2]{\begin{tabular}{@{}#1@{}}#2\end {tabular}}
\caption{The network size and performance for different models of GRD.} \label{tab:II}
\begin{center}
\renewcommand\arraystretch{1.5}
\begin{tabular}{p{2.5cm}<{\centering} p{1.0cm}<{\centering} p{1.0cm}<{\centering}p{0.6cm}<{\centering}p{0.6cm}<{\centering}p{0.6cm}<{\centering}}
\hline\hline
Model & $n_{params}$ & $n_{blocks}$ & $d_{model}$ & $n_{heads}$ & RMSE \\
\hline
2-Link & 141M & 44 & 512 & 8 & 0.180\\

3-Link & 200M & 64 & 512 & 8 & 0.1557\\

\hline
3-Link (Small) & 50M & 16 & 512  & 8 & 0.1862\\

3-Link (Medium) & 100M & 32 & 512  & 8 &  0.1827\\
\hline\hline
\end{tabular}
\end{center}
\end{table}

\begin{table}[t]
\newcommand{\tabincell}[2]{\begin{tabular}{@{}#1@{}}#2\end {tabular}}
\caption{Performance of different inverse dynamics.} \label{tab:III}
\begin{center}
\renewcommand\arraystretch{1.5}
\begin{tabular}{p{4.0cm}<{\centering} p{3.5cm}<{\centering} }
\hline\hline
Model & RMSE  \\
\hline
General inverse dynamics & $4.212 N\cdot m$\\

GRD + general inverse dynamics & $8.127 N\cdot m$ \\

GRD + right inverse model & $2.784 N\cdot m$ \\

General inverse dynamics + GRD & $0.204$ \\

Left inverse model + GRD & $0.076$ \\

\hline\hline
\end{tabular}
\end{center}
\end{table}

\begin{figure*}[t]
\subfigure[]{
\label{fig:inverse_results:a}
\begin{minipage}[b]{0.23\textwidth}
\centering
\includegraphics[scale=0.33]{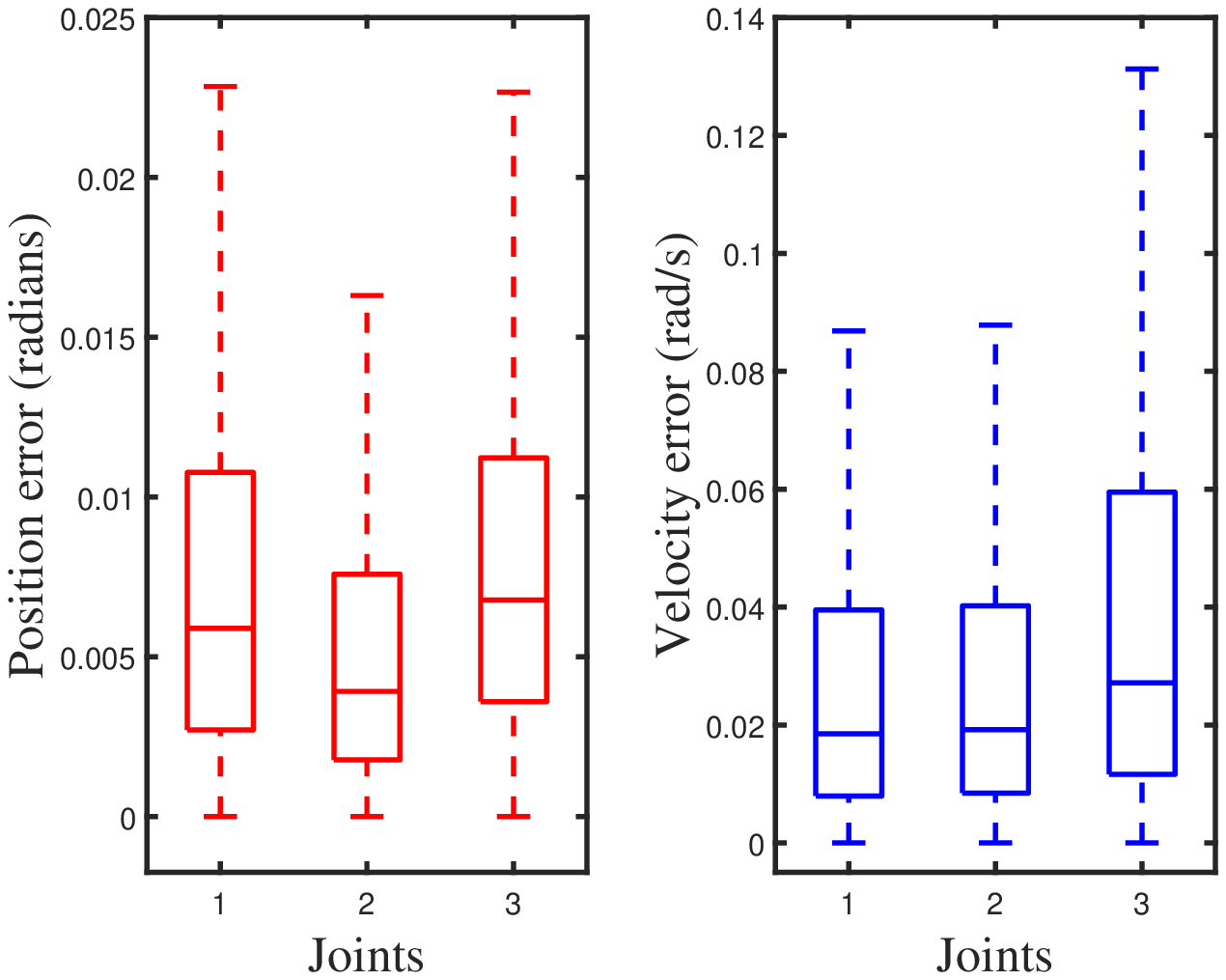}
\end{minipage}}
\subfigure[]{
\label{fig:inverse_results:b}
\begin{minipage}[b]{0.23\textwidth}
\centering
\includegraphics[scale=0.33]{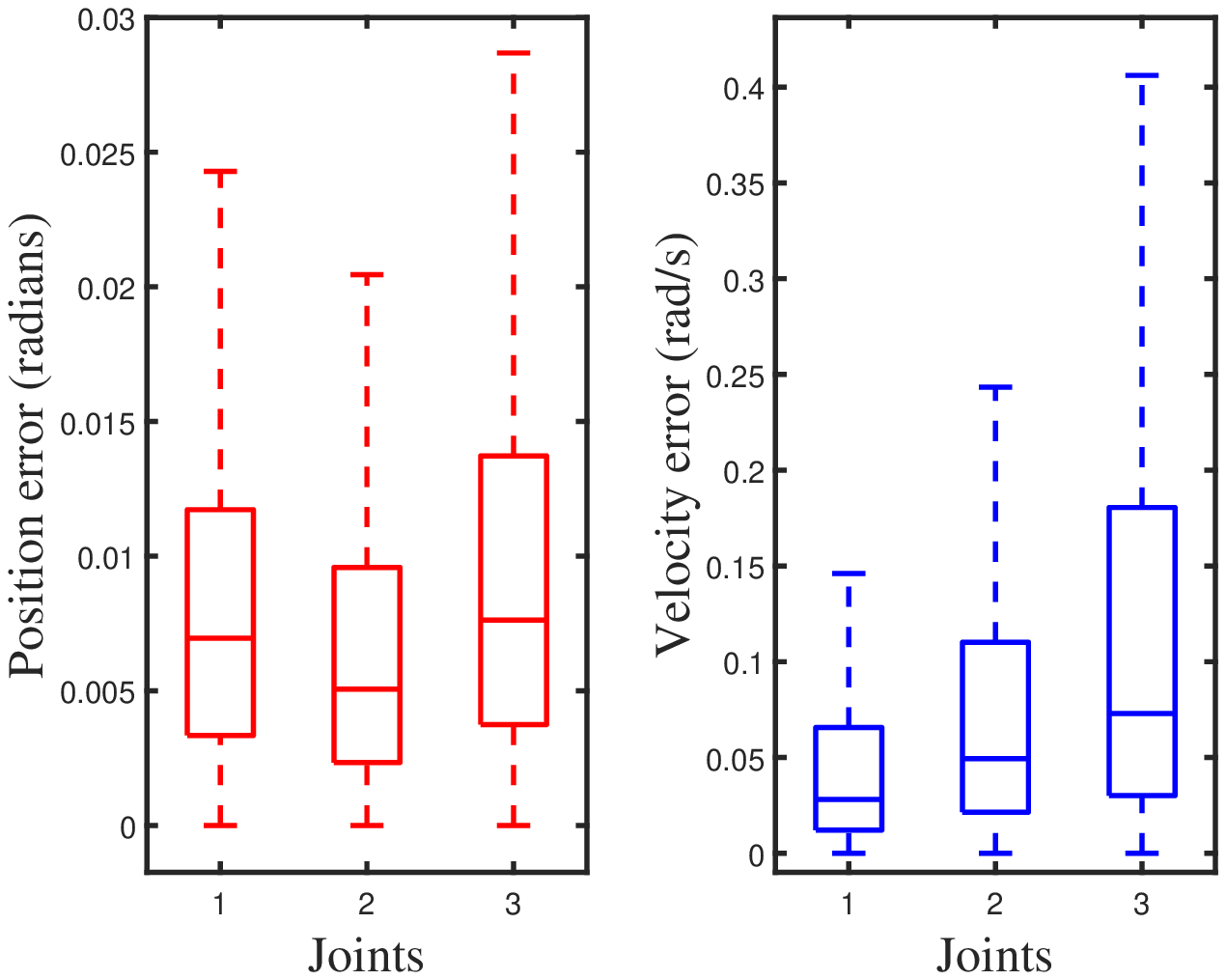}
\end{minipage}}
\subfigure[]{
\label{fig:inverse_results:c}
\begin{minipage}[b]{0.23\textwidth}
\centering
\includegraphics[scale=0.33]{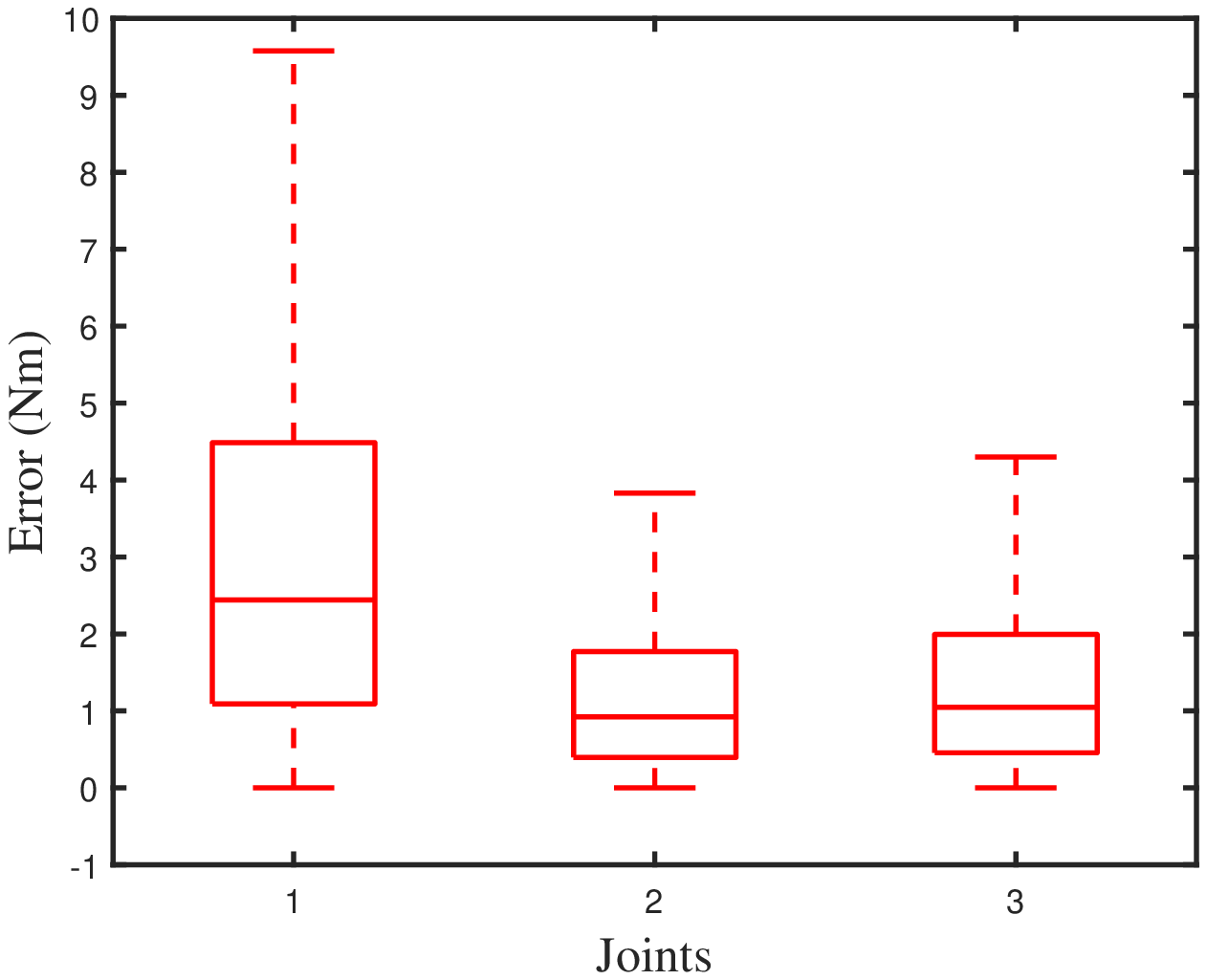}
\end{minipage}}
\subfigure[]{
\label{fig:inverse_results:d}
\begin{minipage}[b]{0.23\textwidth}
\centering
\includegraphics[scale=0.33]{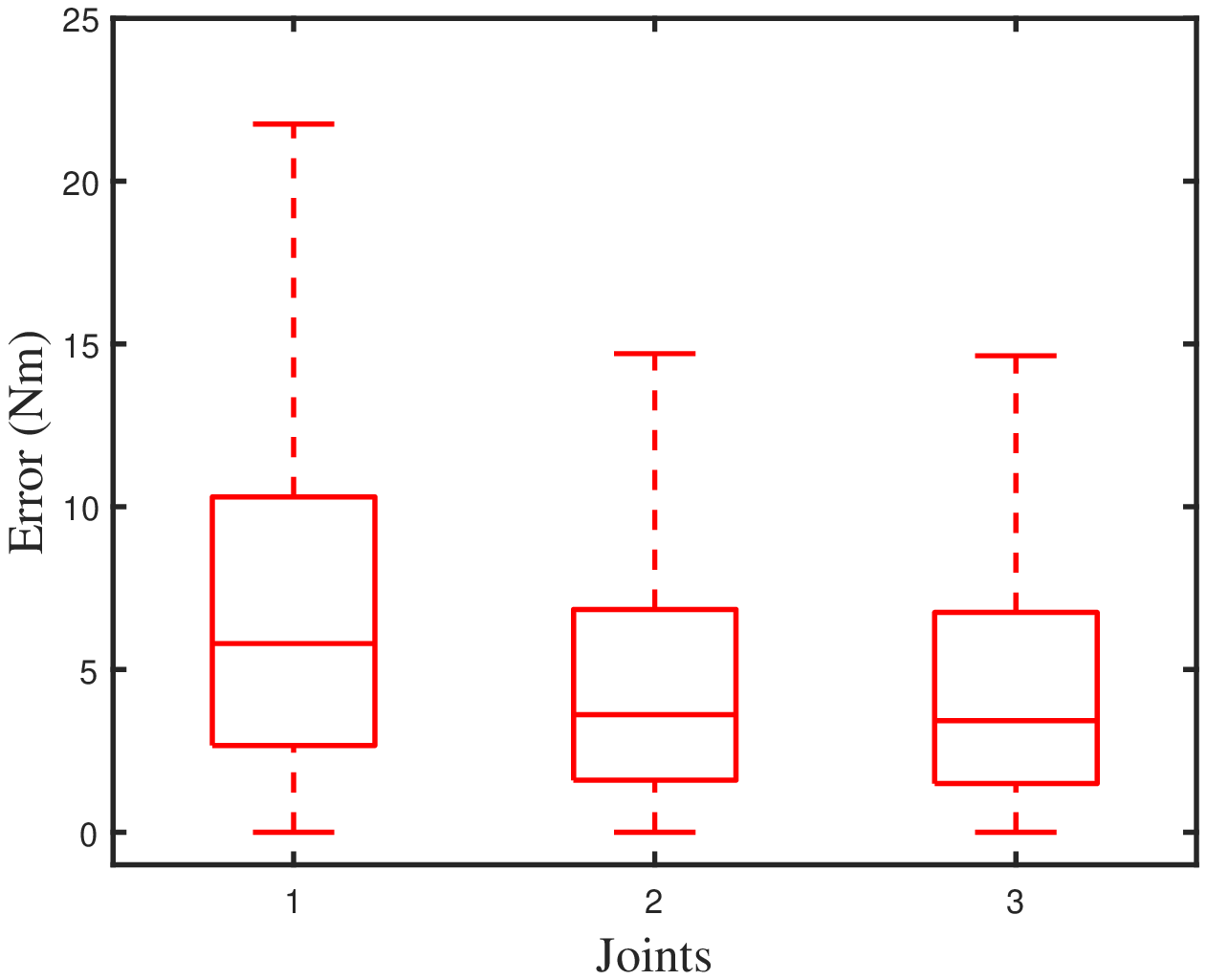}
\end{minipage}}
\caption{The errors of the combination of a) left inverse and GRD, b) general inverse dynamics and GRD, c) GRD and its right inverse, and d) GRD and general inverse dynamics.}
\label{fig:inverse_results}
\end{figure*}

We also display the learning results of inverse models of general dynamics, as shown in Tab. \ref{tab:III}, taking 3-link robots as an example. The three types of inverse models have the same structure size as used in RGD learning. We train the general inverse dynamics and achieve an RMSE of $4.212 N\cdot m$. This inverse dynamics model has a physical meaning and can perform independently. However, if one connects it to the right of GRD, the RMSE is $8.127 N\cdot m$, an almost double error. It shows that, due to model errors, GRD and general inverse dynamics have no apparent physical relationship. We train the right inverse of GRD and obtain an RMSE of $2.784 N\cdot m$, close to one-third error of the combination of general inverse dynamics and GRD. Similar results apply in investigating the left inverse. Putting general inverse dynamics to the left of GRD receives an RMSE of 0.204, about one-third larger than the GRD's error. Combining the left inverse model and GRD reduces the RMSE to 0.076, an approximately one-third error of combining GRD and general inverse dynamics. The above results show that these two inverse models play essential roles in cooperating with GRD, with less accumulated errors, and their combinations are more suitable to instances where dynamics and inverse dynamics are simultaneously needed.

To analyze how those models work with various robots, we generate another one thousand trajectories on a hundred randomly selected models and test the performance of connecting GRD with different inverse models. Fig. \ref{fig:inverse_results} shows the results of the error distribution of four different combinations. Since the state includes position and velocity, we exhibit them separately in the first two subfigures. Fig. \ref{fig:inverse_results:a} shows the results of combining left inverse and GRD. The mean of each joint position error is about 0.006 $radians$, a tiny deviation, and the average of the joint velocity errors is about 0.02 $rad/s$. We can conclude that the position is more precise than the velocity using GRD and its inverse. Fig. \ref{fig:inverse_results:b} replaces the left inverse with general inverse dynamics, and their results show that the position errors demonstrate little difference, but a substantial divergence exists between velocity errors. It appears that the training of the left inverse model from general inverse dynamics primarily reduces the error of joint velocities. Fig. \ref{fig:inverse_results:c} presents the torque results of the combination of GRD and its right inverse. The base joint exerts a mean of 2.5 $N\cdot m$ torque errors, and the error means of the other two joints are around 1 $N\cdot m$. Replacing the right inverse with general inverse dynamics leads to torque error enlargement of every joint. These results demonstrate that the two inverse models present fewer errors in joint velocities and torques than general inverse dynamics when connecting to GRD.

To manifest the performance in learning various inverse dynamics, we also compare with LSTM and linear network, using the trajectories of 2-link and 3-link robots, as tested in Fig. \ref{fig:dynamics_result:a}. Fig \ref{fig:dynamics_result:b} displays the error distribution of different methods, where the linear network has the largest RMSE error, and our general model still performs the best. It shows the superiority of our inverse dynamics models.

\subsection{Transfer to reality}

\begin{figure}[t]
 \centerline{\psfig{file=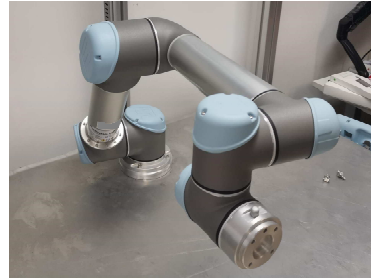,scale=0.4}}
 \caption{The UR5e robot used in experiments.} \label{fig:rob}
\end{figure}

\begin{figure*}[t]
\subfigure[]{
\label{fig:transfer_results:a}
\begin{minipage}[b]{0.183\textwidth}
\centering
\includegraphics[scale=0.27]{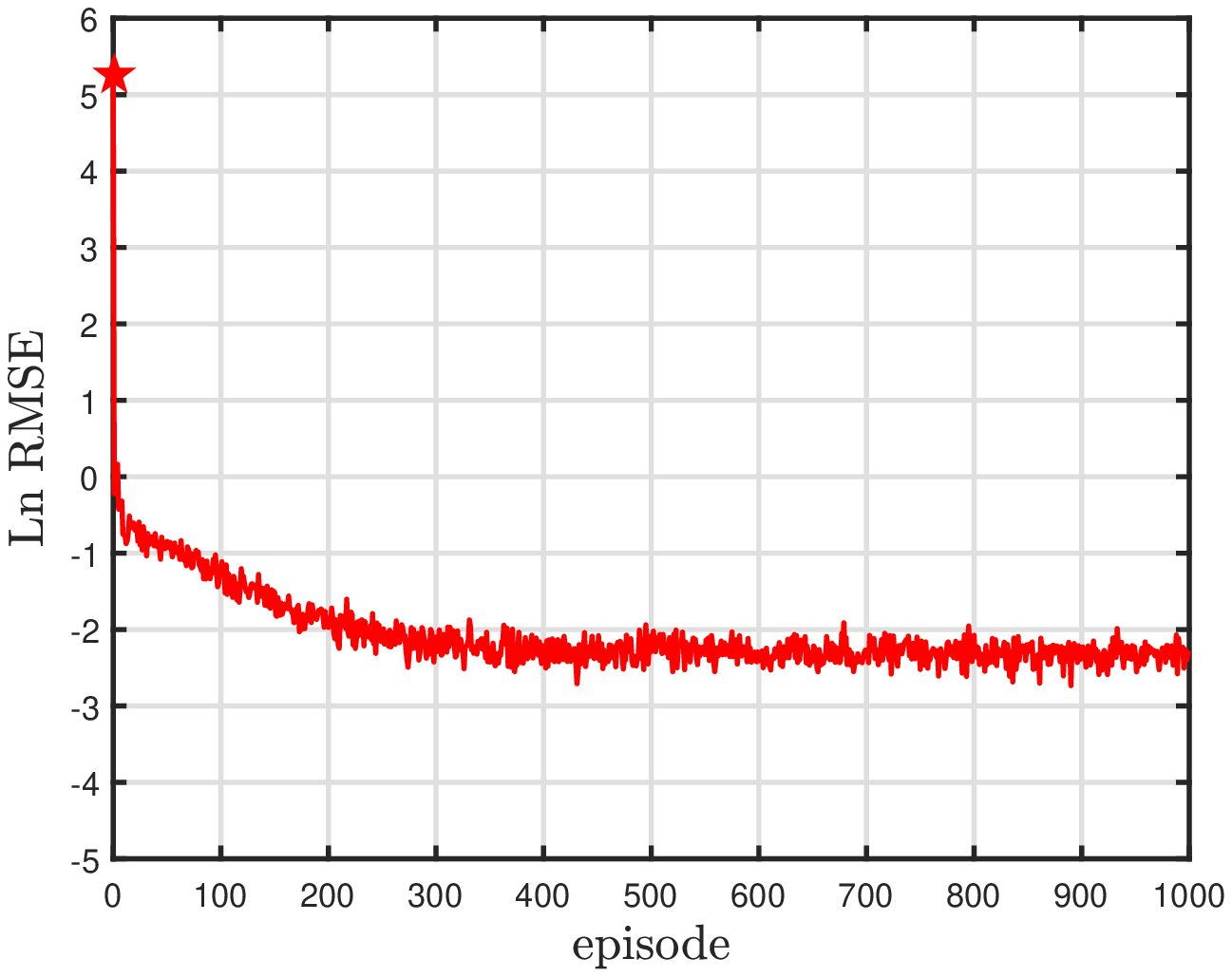}
\end{minipage}}
\subfigure[]{
\label{fig:transfer_results:b}
\begin{minipage}[b]{0.183\textwidth}
\centering
\includegraphics[scale=0.27]{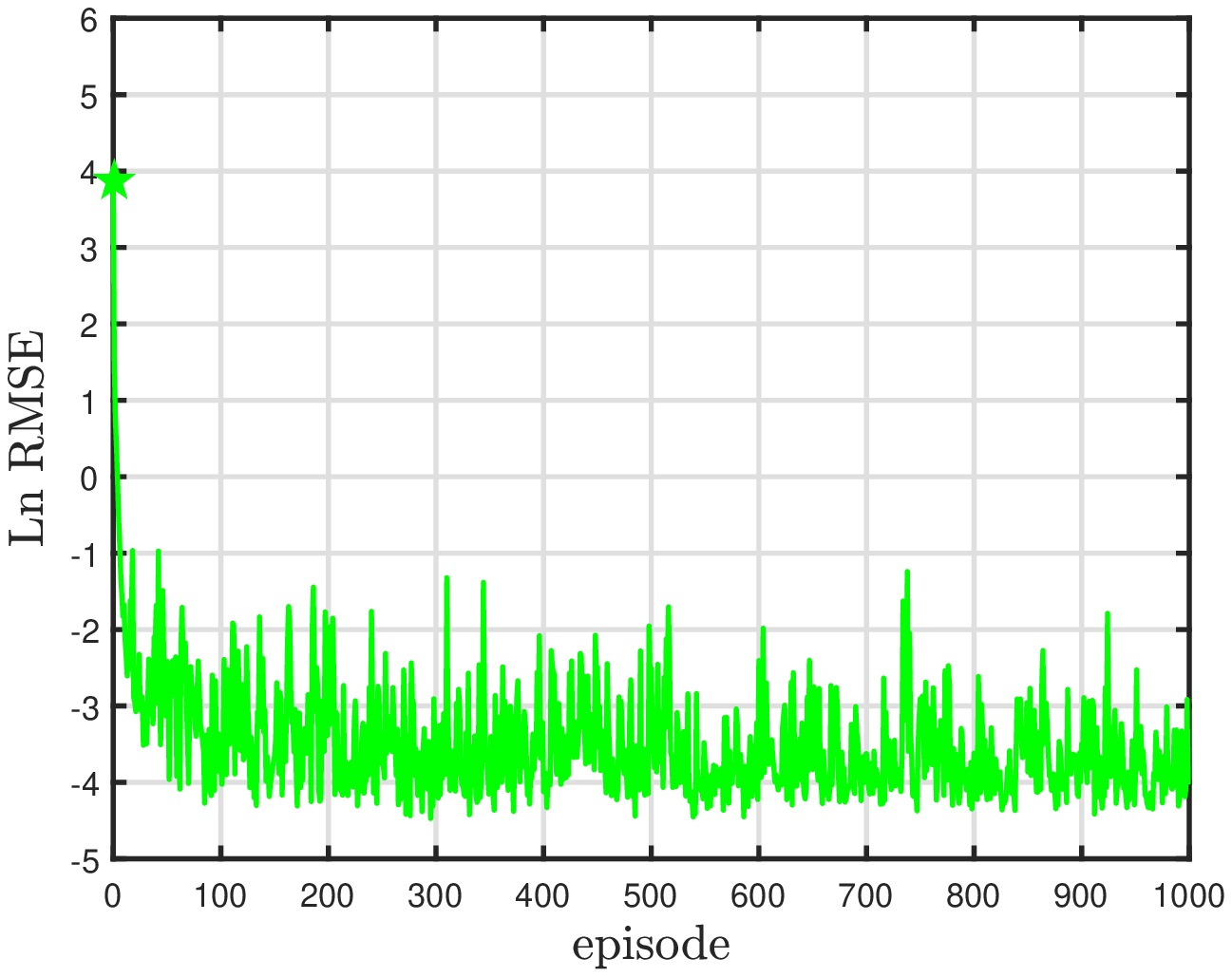}
\end{minipage}}
\subfigure[]{
\label{fig:transfer_results:c}
\begin{minipage}[b]{0.183\textwidth}
\centering
\includegraphics[scale=0.27]{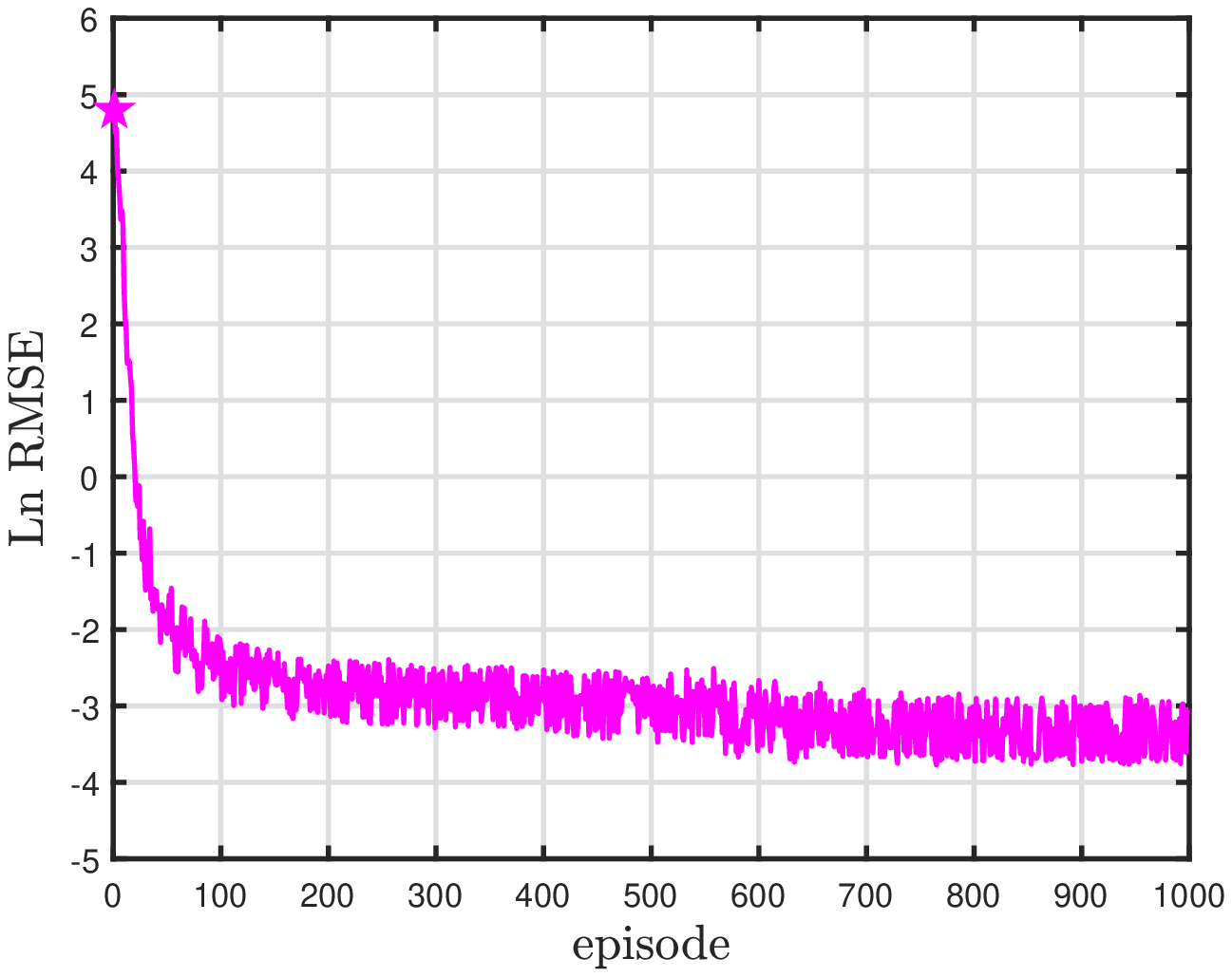}
\end{minipage}}
\subfigure[]{
\label{fig:transfer_results:d}
\begin{minipage}[b]{0.183\textwidth}
\centering
\includegraphics[scale=0.27]{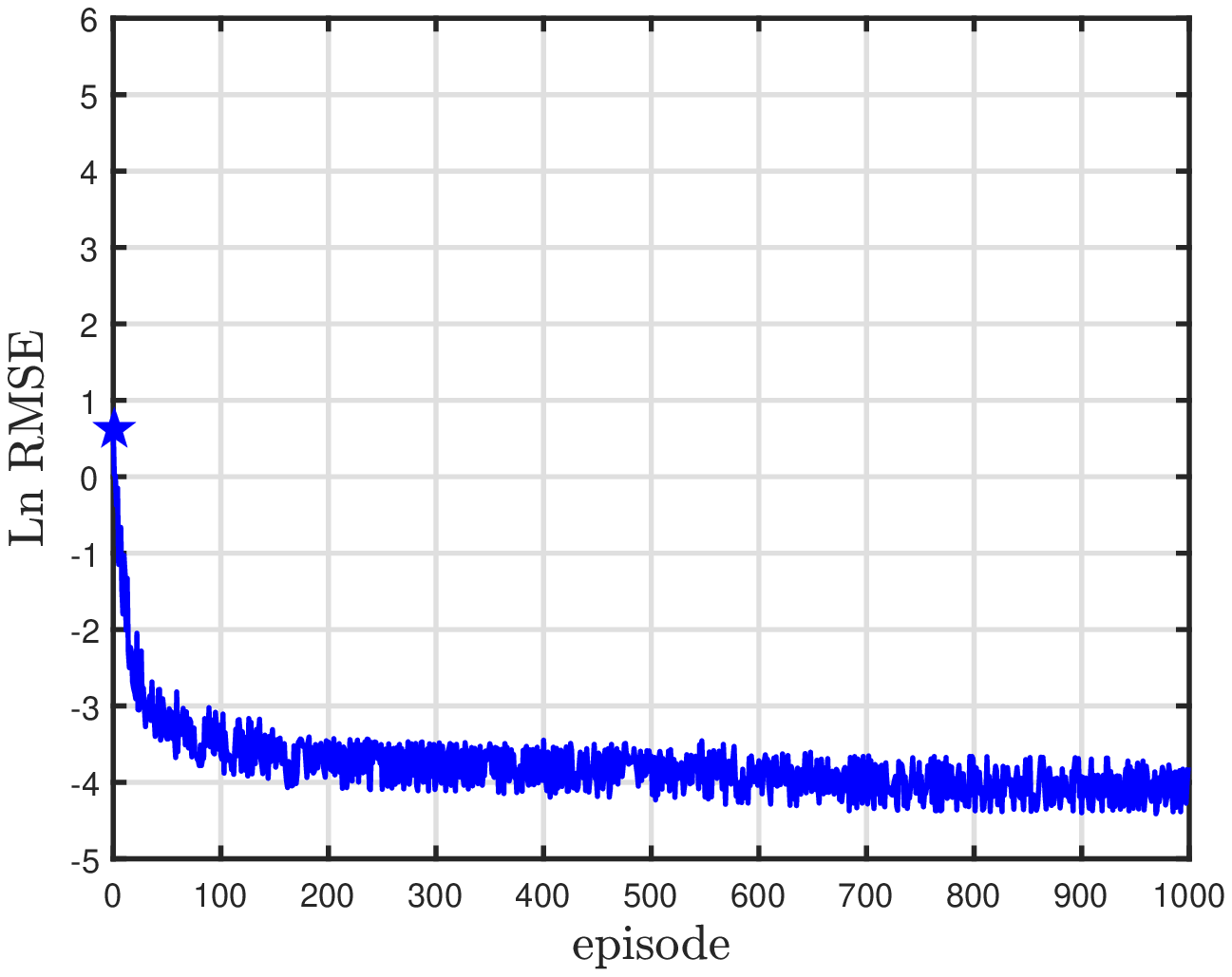}
\end{minipage}}
\subfigure[]{
\label{fig:transfer_results:e}
\begin{minipage}[b]{0.183\textwidth}
\centering
\includegraphics[scale=0.27]{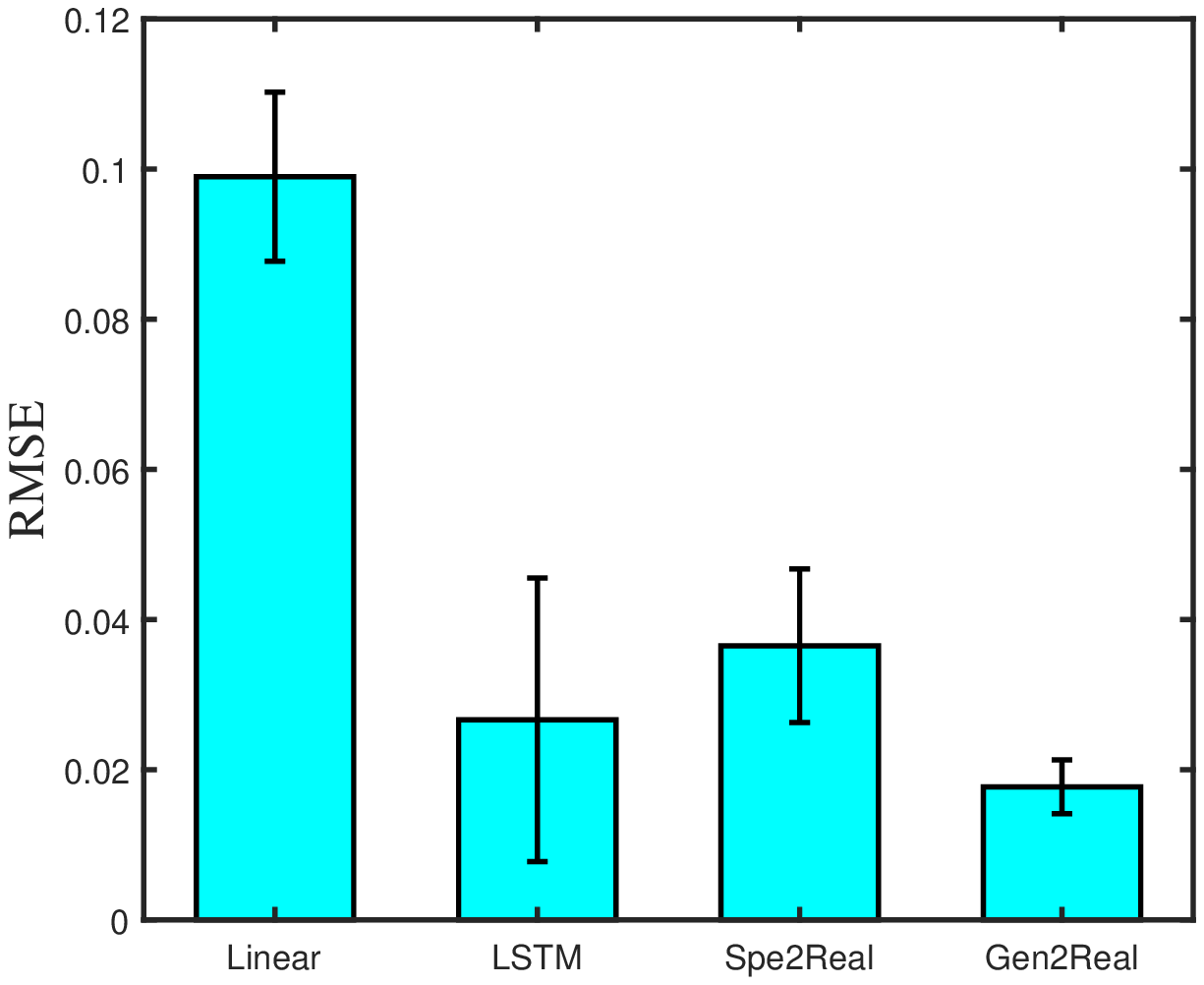}
\end{minipage}}\\
\subfigure[]{
\label{fig:transfer_results:f}
\begin{minipage}[b]{0.183\textwidth}
\centering
\includegraphics[scale=0.27]{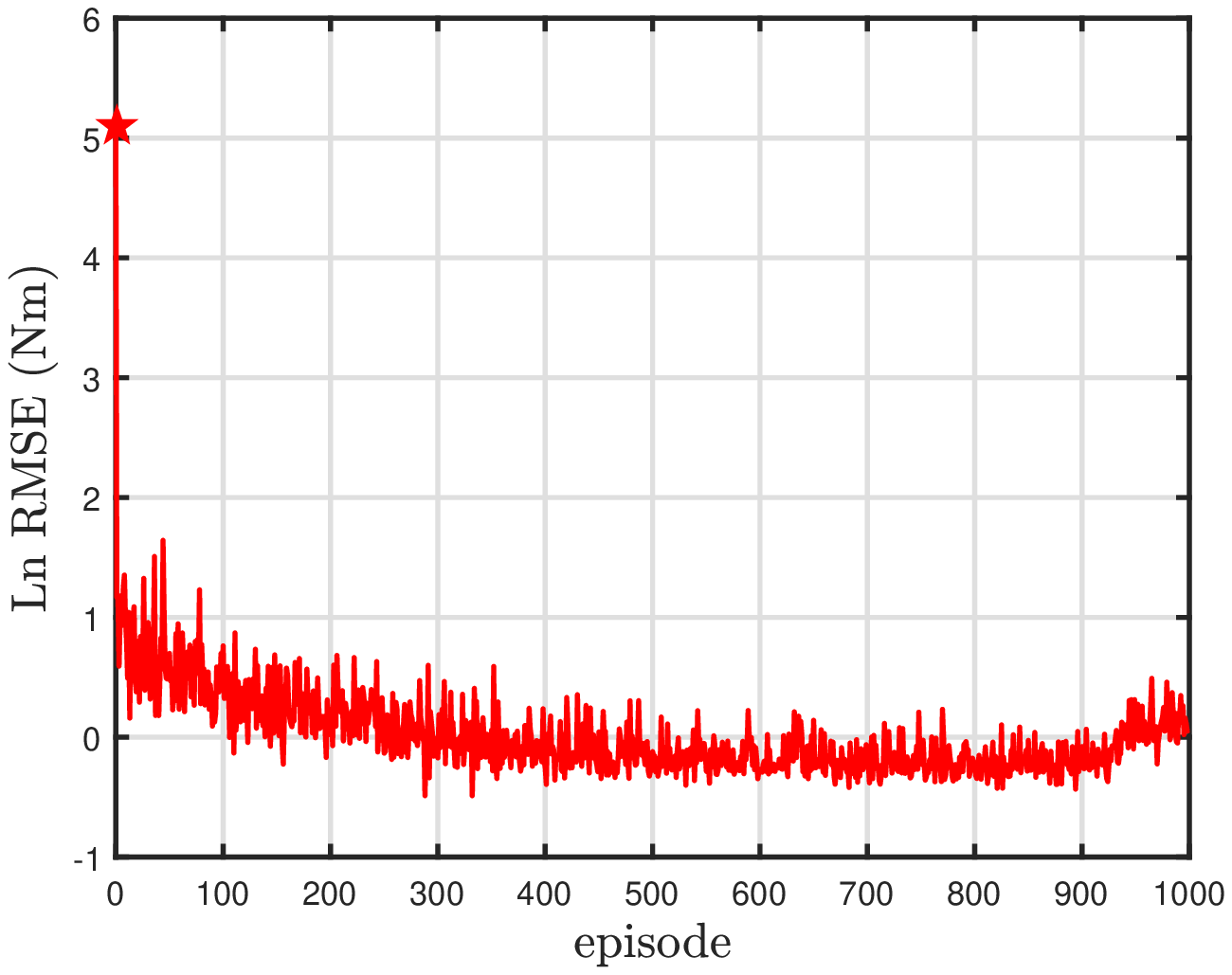}
\end{minipage}}
\subfigure[]{
\label{fig:transfer_results:g}
\begin{minipage}[b]{0.183\textwidth}
\centering
\includegraphics[scale=0.27]{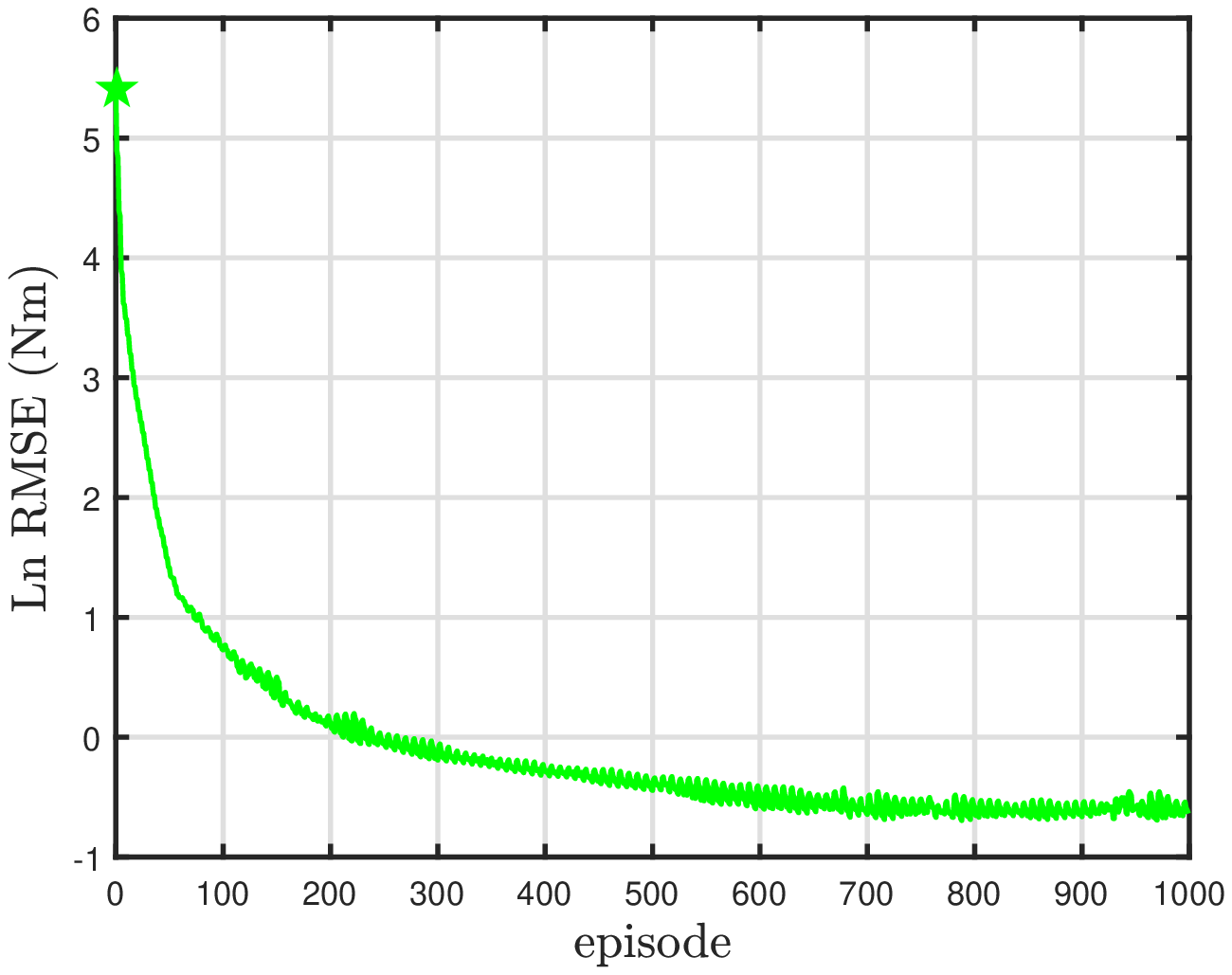}
\end{minipage}}
\subfigure[]{
\label{fig:transfer_results:h}
\begin{minipage}[b]{0.183\textwidth}
\centering
\includegraphics[scale=0.27]{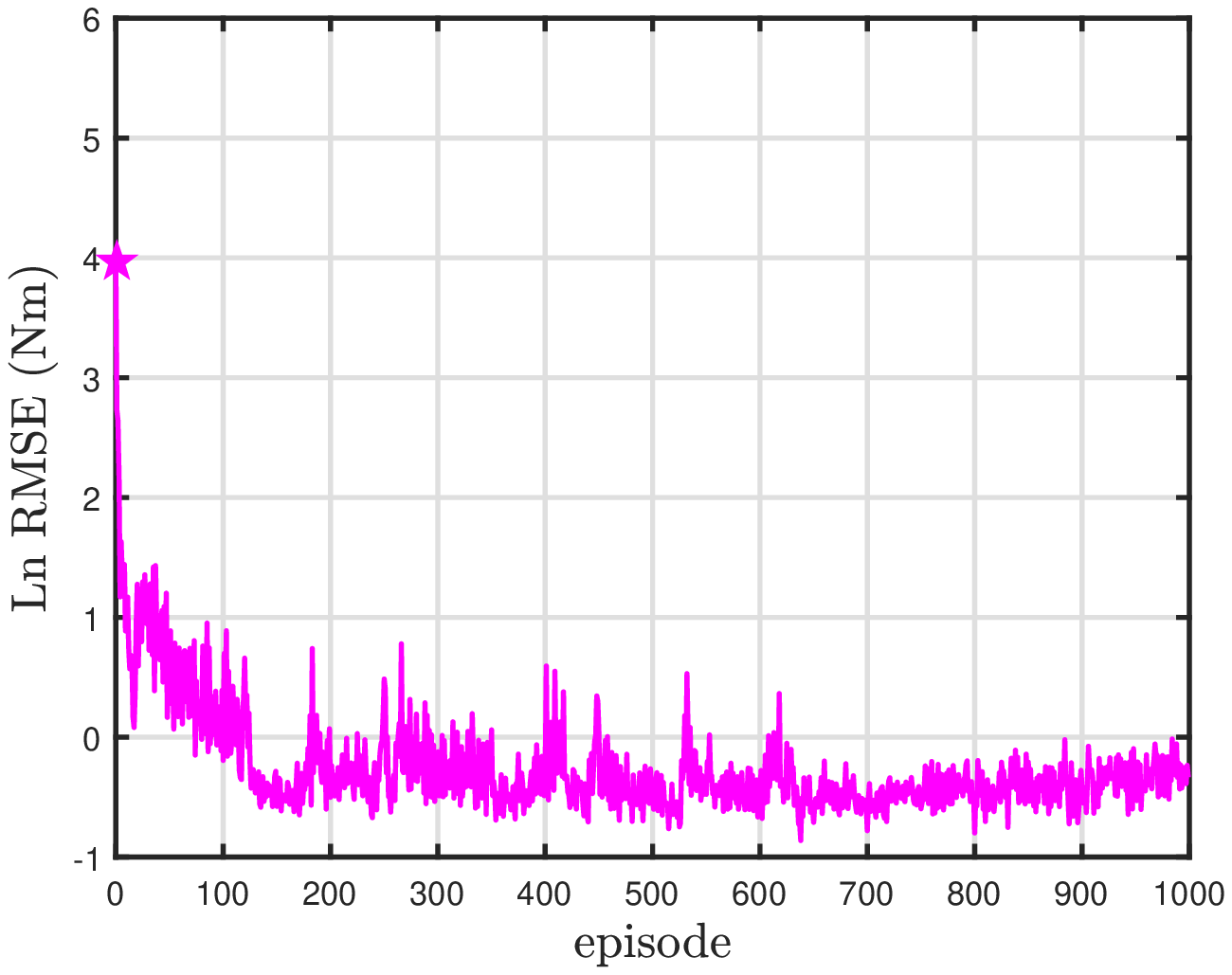}
\end{minipage}}
\subfigure[]{
\label{fig:transfer_results:i}
\begin{minipage}[b]{0.183\textwidth}
\centering
\includegraphics[scale=0.27]{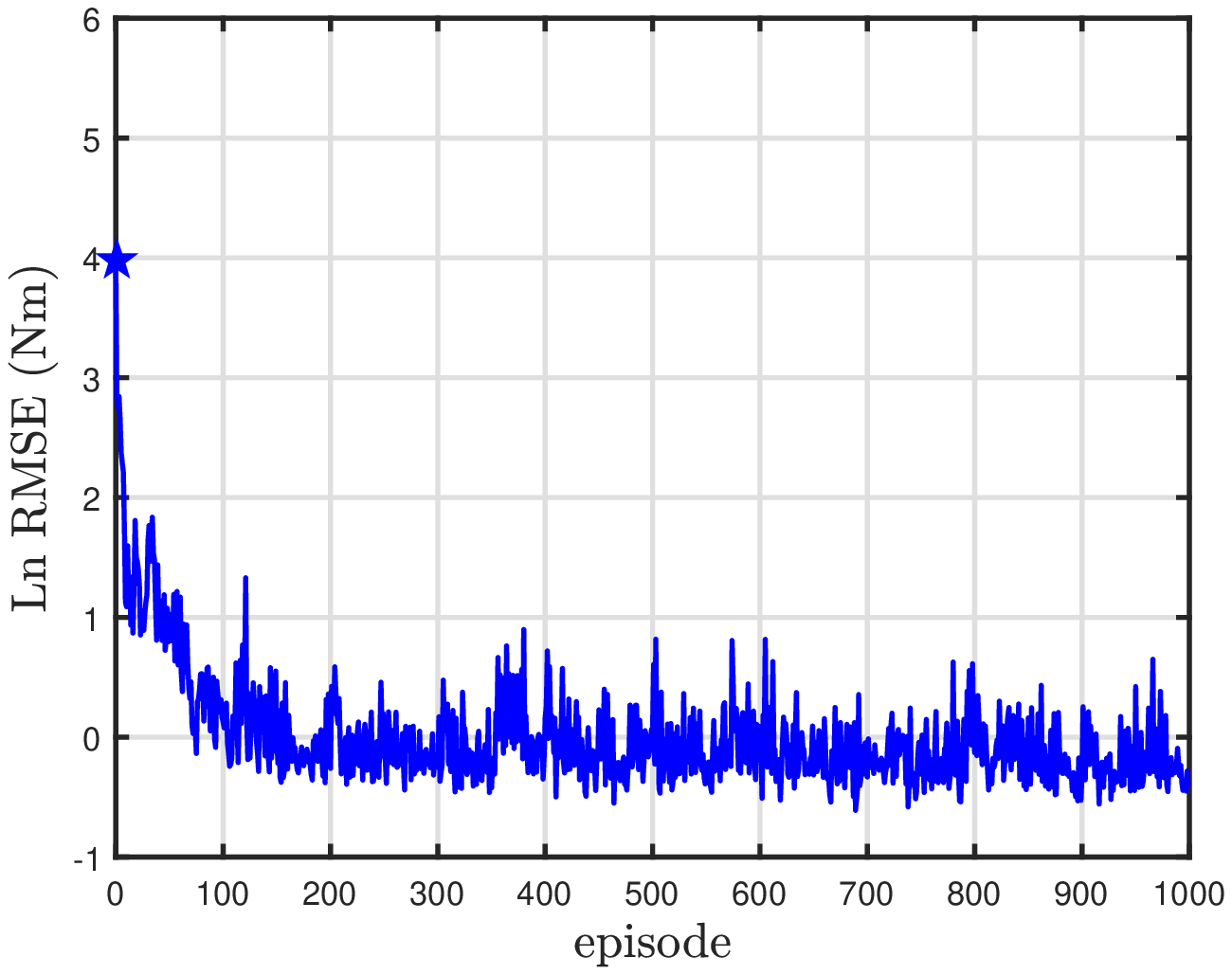}
\end{minipage}}
\subfigure[]{
\label{fig:transfer_results:j}
\begin{minipage}[b]{0.183\textwidth}
\centering
\includegraphics[scale=0.27]{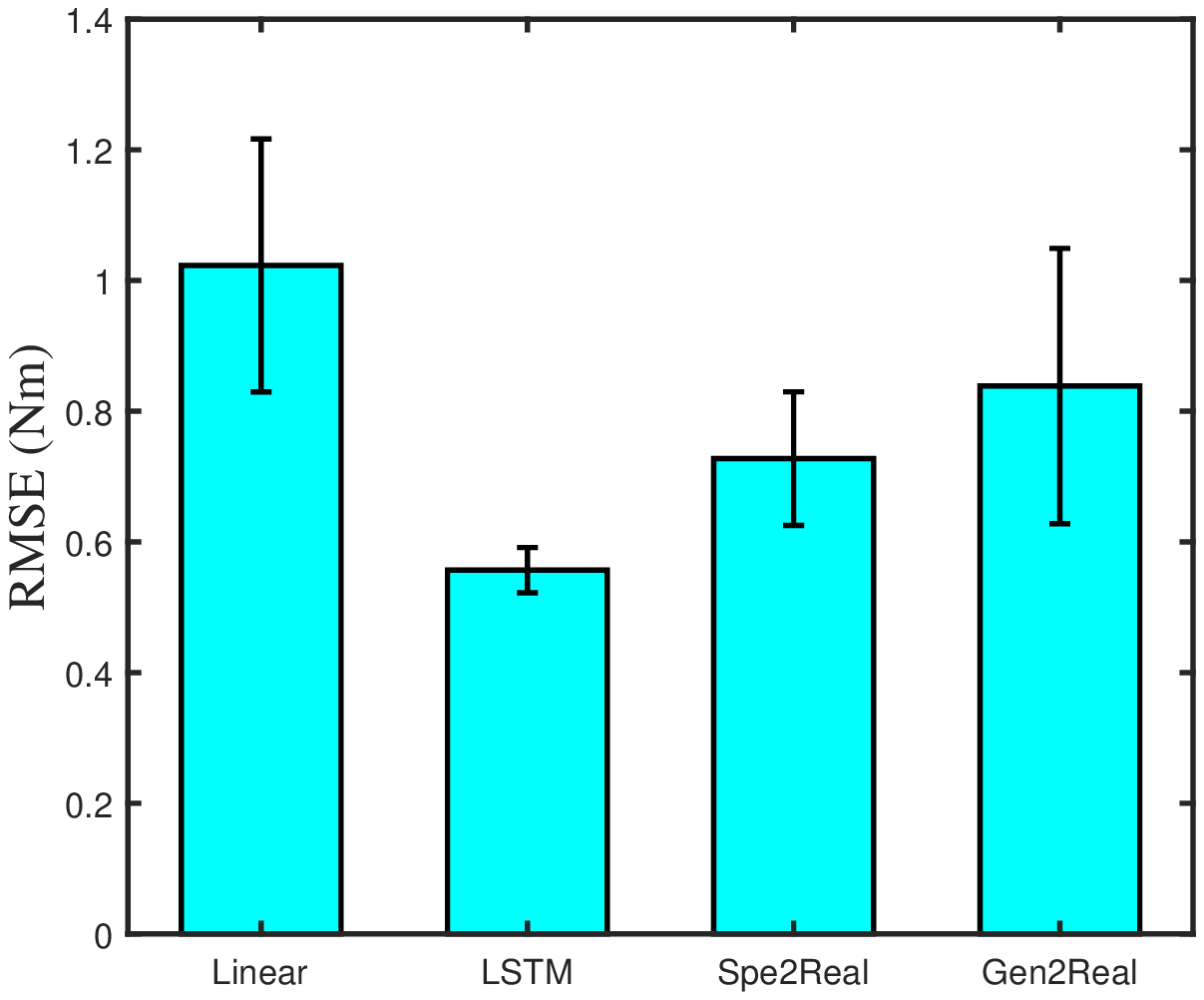}
\end{minipage}}
\caption{The training results of different methods in transferring to reality, where log of RMSE is used. a)-d) use a linear network, LSTM, Spe2Real, and Gen2Real to transfer dynamics models to reality. e) is the error distributions of the last 100 steps. f)-i) use the four models to transfer inverse dynamics models to reality. j) is their error distributions of the last 100 steps.}
\label{fig:transfer_results}
\end{figure*}

The UR5e robot, as shown in Fig. \ref{fig:rob}, applies in transfer validation. We run the robot with motor bubbling and collect a trajectory consisting of five thousand time steps, in which $80\%$ is for training and the rest for testing. In transferring to reality, we only adjust the last layer of our network since it has a massive scale. As for comparisons, we also employ the two comparative methods, LSTM and linear network, with the same structure used in the previous subsection, for Sim2Real. We use a similar parameter range, $[0.25, 4]$, as used in the document \cite{Peng}, and train these two networks.

Fig. \ref{fig:transfer_results} shows the transferring results of different methods in a thousand episodes, where the star means the error of directly applying to experiment data without transferring, which displays that the general model is more compatible with the unseen robots. All models approximate the physical dynamics with acceptable learning errors, with the linear network having the largest errors. The upper half of Fig. \ref{fig:transfer_results} shows the result of dynamics transfer. Gen2Real has the best performance, with a mean of 0.018 and a standard deviation of 0.004. Spe2Real exhibits a moderate error, with a mean of 0.036 and a standard deviation of 0.010. The lower half of Fig. \ref{fig:transfer_results} is the result of transferring inverse dynamics. LSTM is the best, and Spe2Real is close, with an RMSE mean of 0.72 $N\cdot m$ and a standard deviation of 0.10 $N\cdot m$. Gen2Real is relatively larger, with a mean of 0.83 $N\cdot m$ and a standard deviation of 0.21 $N\cdot m$. These results reveal that Gen2Real and Spe2Real can transfer the general models to reality with competitive performance.

\begin{figure}[t]
\subfigure[]{
\label{fig:another_results:a}
\begin{minipage}[b]{0.23\textwidth}
\centering
\includegraphics[scale=0.33]{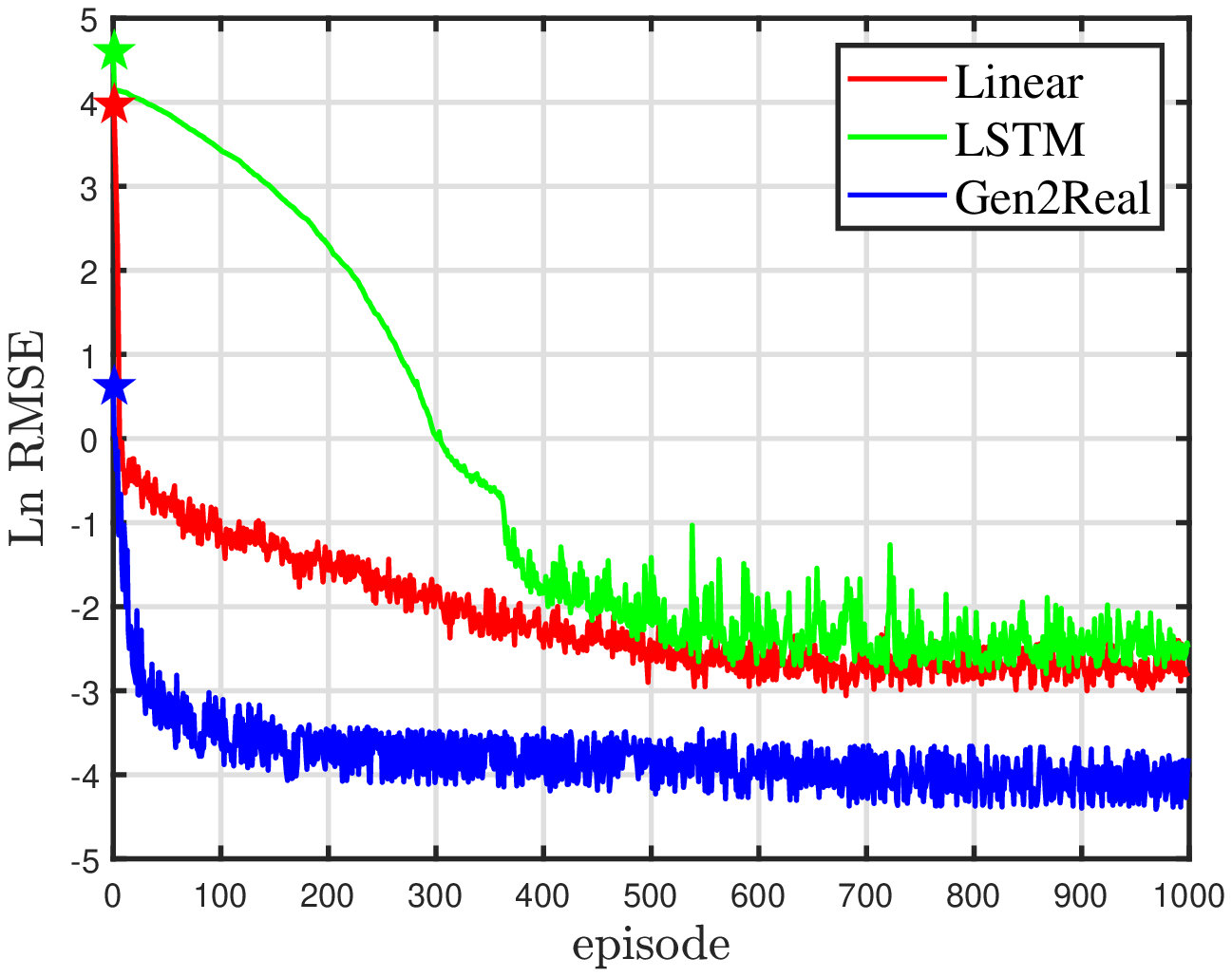}
\end{minipage}}
\subfigure[]{
\label{fig:another_results:b}
\begin{minipage}[b]{0.23\textwidth}
\centering
\includegraphics[scale=0.33]{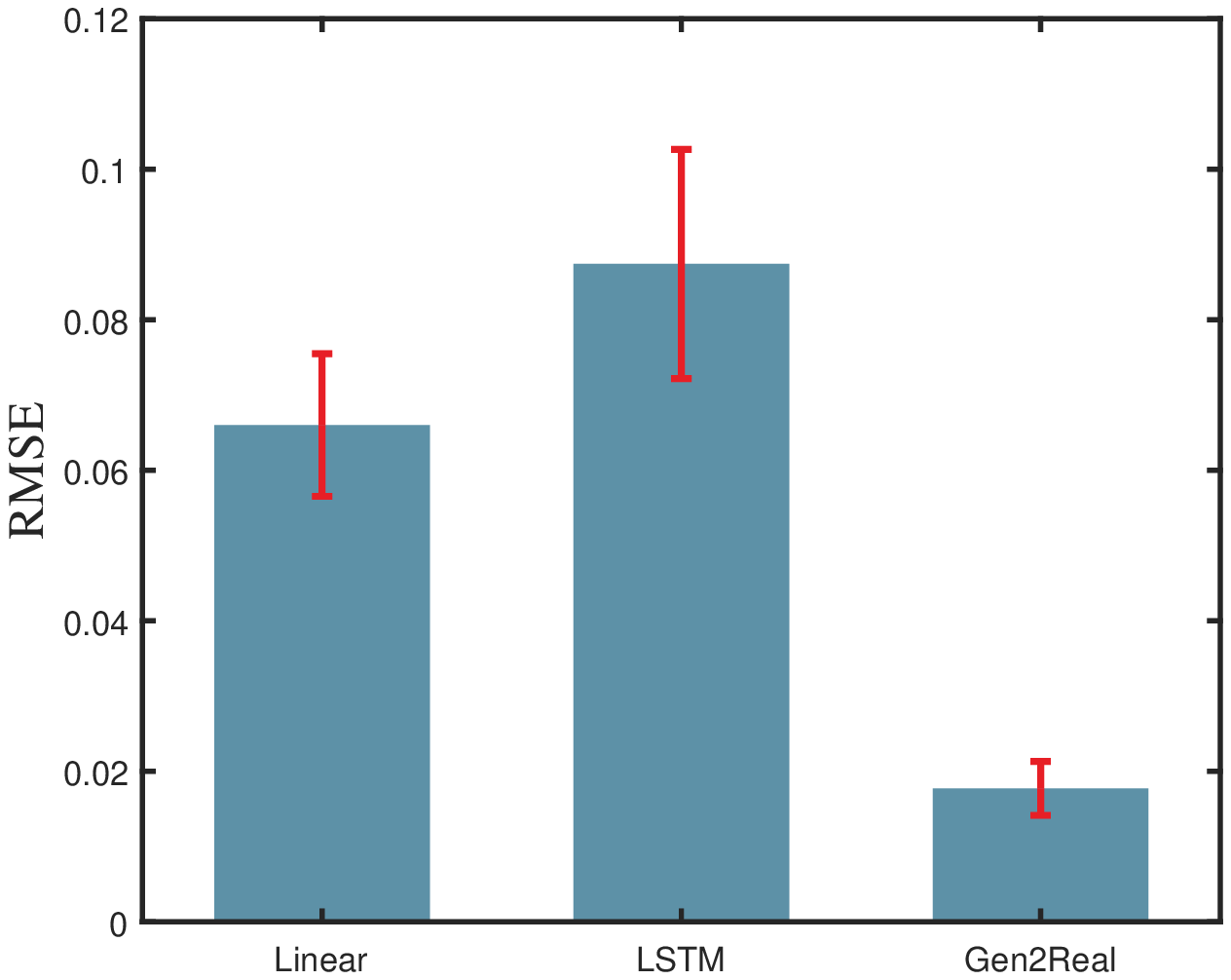}
\end{minipage}}
\caption{The results of various methods in learning a different robot dynamics and transferring to UR5. They learn a 2-link model with randomized parameters in simulation and transfer to UR5 in reality. a) The learning process of transfer. b) The distribution of RMSEs of the last 100 steps.}
\label{fig:another_results}
\end{figure}

We test the capability of transferring to another robot set, which is entirely different from the UR5e model. We generate a new dataset by randomizing a 2-link robot with $[0.25, 4]$ parameter range and different joint axes. We train the comparative methods with the data and then transfer them to the UR5 robot. Fig. \ref{fig:another_results} shows the results. It takes LSTM a long time to converge, with a mean of 5 at the end. The linear network's behavior is similar to its Sim2Real. Compare with them, Gen2Real has outstanding performance, with a one-fourth RMSE mean and a one-third standard variation. This comparison validates that our models are superior in a different robot set transfer.

\begin{figure}[t]
\subfigure[]{
\label{fig:another_results:a}
\begin{minipage}[b]{0.23\textwidth}
\centering
\includegraphics[scale=0.33]{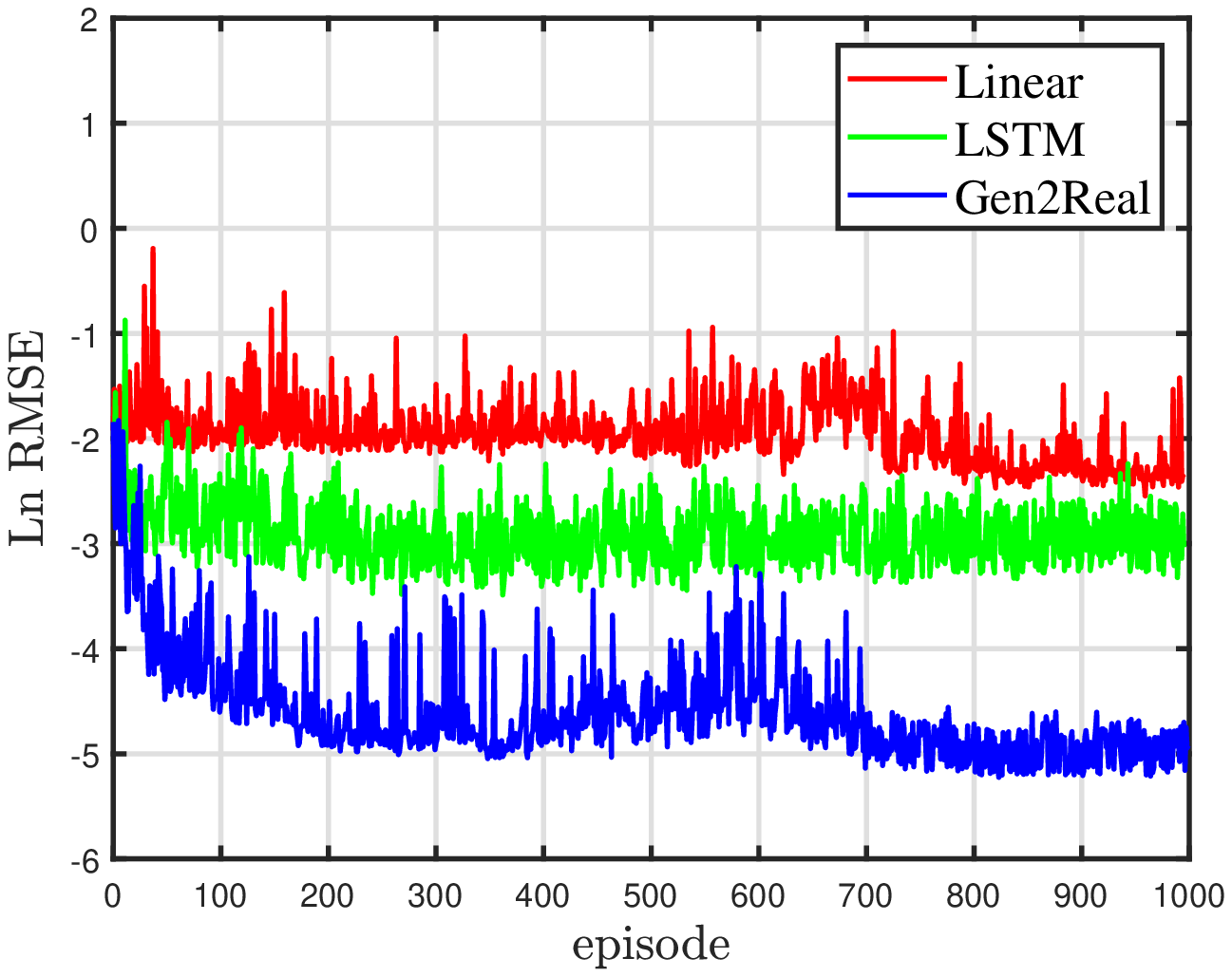}
\end{minipage}}
\subfigure[]{
\label{fig:another_results:b}
\begin{minipage}[b]{0.23\textwidth}
\centering
\includegraphics[scale=0.33]{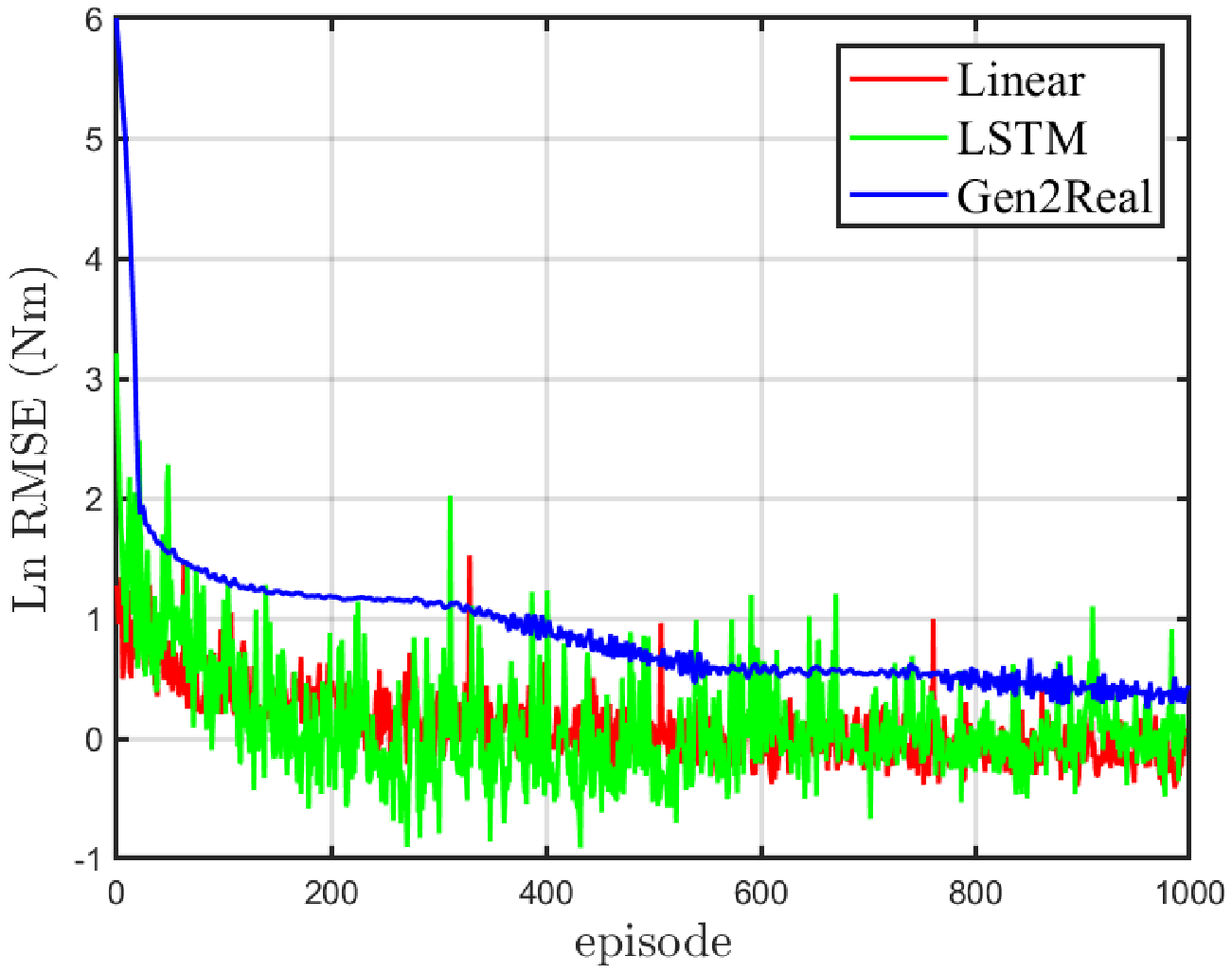}
\end{minipage}}
\caption{The transfer results of different methods in transferring a combination of robot dynamics and its inverse. a) Dynamics follows its inverse. b) Dynamics is followed by its inverse.}
\label{fig:DDtransfer_results}
\end{figure}

We also transfer the combination of robot dynamics and its inverse to reality. Fig. \ref{fig:DDtransfer_results} shows the results of different methods. For the connection of dynamics and its inverse, Gen2Real outperforms the other two methods, with distinct differences in approximation error. For the sequence that dynamics follows its inverse, Gen2Real is inferior to others, but their RMSEs are not much different.

In summary, our general models and Gen2Real have exhibited outstanding performance. In modeling various robots in simulation, GRD is superior to the comparative methods. In transferring to reality, Gen2Real is predominant for dynamics transfer and presents a competent performance for inverse dynamics. All in all, the ``generality'' of our models is revealed.

\section{Conclusion}
We study the learning of general robot dynamics and its inverse models for the first time. These models can promote robot policy learning since they incorporate the dynamics of massive robots. We consider the ``generality'' from the viewpoint of different robot properties. In the order of influence on dynamics, dynamics parameters change the attribute of robot links; topology configurations change the type of robot joints; model dimensions change the number of robot links. We generate a dataset by randomizing these three elements for various robot models and driving torques for motion trajectory. We employ a structure by modifying GPT to adapt to learn general dynamics. We further propose and train three inverse models of GRD, each for specific applications. The results display the advantages of the proposed models, outbalancing other methods in approximating various robot dynamics.

We also investigate Gen2Real to transfer general models to reality. This method can save some tedious processes in Sim2Real and thereby lower the threshold of dynamics learning. Comparisons with other methods show that Gen2Real achieves moderate performance in transferring to reality and superior results to a different robot set.




\begin{thebibliography}{38}
\providecommand{\natexlab}[1]{#1}
\providecommand{\url}[1]{\texttt{#1}}
\expandafter\ifx\csname urlstyle\endcsname\relax
  \providecommand{\doi}[1]{doi: #1}\else
  \providecommand{\doi}{doi: \begingroup \urlstyle{rm}\Url}\fi

\bibitem[Andrychowicz et~al.(2020)Andrychowicz, Baker, Chociej, Jozefowicz,
  McGrew, Pachocki, Petron, Plappert, Powell, Ray, Schneider, Sidor, Tobin,
  Welinder, Weng, and Zaremba]{Andrychowicz}
Marcin Andrychowicz, Bowen Baker, Maciek Chociej, Rafal Jozefowicz, Bob McGrew,
  Jakub Pachocki, Arthur Petron, Matthias Plappert, Glenn Powell, Alex Ray,
  Jonas Schneider, Szymon Sidor, Josh Tobin, Peter Welinder, Lilian Weng, and
  Wojciech Zaremba.
\newblock Learning dexterous in-hand manipulation.
\newblock \emph{The International Journal of Robotics Research}, 39\penalty0
  (1):\penalty0 3--20, 2020.

\bibitem[{Bird} et~al.(preprinted){Bird}, {Pritchard}, {Fratini}, {Ekart}, and
  {Faria}]{Bird}
Jordan~J. {Bird}, Michael~George {Pritchard}, Antonio {Fratini}, Aniko {Ekart},
  and Diego {Faria}.
\newblock Synthetic biological signals machine-generated by gpt-2 improve the
  classification of eeg and emg through data augmentation.
\newblock \emph{IEEE Robotics and Automation Letters}, preprinted.

\bibitem[Brown et~al.(2020{\natexlab{a}})Brown, Mann, Ryder, Subbiah, Kaplan,
  Dhariwal, Neelakantan, Shyam, Sastry, Askell, Agarwal, Herbert-Voss, Krueger,
  Henighan, Child, Ramesh, Ziegler, Wu, Winter, Hesse, Chen, Sigler, Litwin,
  Gray, Chess, Clark, Berner, McCandlish, Radford, Sutskever, and
  Amodei]{Brown}
Tom~B Brown, Ben Mann, Nick Ryder, Melanie Subbiah, Jared~D Kaplan, Prafulla
  Dhariwal, Arvind Neelakantan, Pranav Shyam, Girish Sastry, Amanda Askell,
  Sandhini Agarwal, Ariel Herbert-Voss, Gretchen~M Krueger, Tom Henighan, Rewon
  Child, Aditya Ramesh, Daniel Ziegler, Jeffrey Wu, Clemens Winter, Chris
  Hesse, Mark Chen, Eric Sigler, Mateusz Litwin, Scott Gray, Benjamin Chess,
  Jack Clark, Christopher Berner, Sam McCandlish, Alec Radford, Ilya Sutskever,
  and Dario Amodei.
\newblock Language models are few-shot learners.
\newblock In \emph{Neural Information Processing Systems}, 2020{\natexlab{a}}.

\bibitem[Brown et~al.(2020{\natexlab{b}})Brown, Mann, Ryder, Subbiah, Kaplan,
  Dhariwal, Neelakantan, Shyam, Sastry, Askell, Agarwal, Herbert-Voss, Krueger,
  Henighan, Child, Ramesh, Ziegler, Wu, Winter, Hesse, Chen, Sigler, Litwin,
  Gray, Chess, Clark, Berner, McCandlish, Radford, Sutskever, and
  Amodei]{brown2020language}
Tom~B. Brown, Benjamin Mann, Nick Ryder, Melanie Subbiah, Jared Kaplan,
  Prafulla Dhariwal, Arvind Neelakantan, Pranav Shyam, Girish Sastry, Amanda
  Askell, Sandhini Agarwal, Ariel Herbert-Voss, Gretchen Krueger, Tom Henighan,
  Rewon Child, Aditya Ramesh, Daniel~M. Ziegler, Jeffrey Wu, Clemens Winter,
  Christopher Hesse, Mark Chen, Eric Sigler, Mateusz Litwin, Scott Gray,
  Benjamin Chess, Jack Clark, Christopher Berner, Sam McCandlish, Alec Radford,
  Ilya Sutskever, and Dario Amodei.
\newblock Language models are few-shot learners, 2020{\natexlab{b}}.

\bibitem[Chebotar et~al.(2019)Chebotar, Handa, Makoviychuk, Macklin, Issac,
  Ratliff, and Fox]{Chebotar}
Yevgen Chebotar, Ankur Handa, Viktor Makoviychuk, Miles Macklin, Jan Issac,
  Nathan Ratliff, and Dieter Fox.
\newblock Closing the sim-to-real loop: Adapting simulation randomization with
  real world experience.
\newblock In \emph{International Conference on Robotics and Automation}, pages
  8973--8979, 2019.

\bibitem[Chen et~al.(2020)Chen, Radford, Child, Wu, Jun, Luan, and
  Sutskever]{Chen}
Mark Chen, Alec Radford, Rewon Child, Jeffrey Wu, Heewoo Jun, David Luan, and
  Ilya Sutskever.
\newblock Generative pretraining from pixels.
\newblock In \emph{International Conference on Machine Learning}, pages
  1691--1703, 2020.

\bibitem[Chung and Glass(2020)]{Chung}
Yu-An Chung and James Glass.
\newblock Generative pre-training for speech with autoregressive predictive
  coding.
\newblock In \emph{IEEE International Conference on Acoustics, Speech and
  Signal Processing}, pages 3497--3501, 2020.

\bibitem[{Desai} et~al.(2020){Desai}, {Karnan}, {Hanna}, {Warnell}, and
  {Stone}]{Desai}
S.~{Desai}, H.~{Karnan}, J.~P. {Hanna}, G.~{Warnell}, and P.~{Stone}.
\newblock Stochastic grounded action transformation for robot learning in
  simulation.
\newblock In \emph{IEEE/RSJ International Conference on Intelligent Robots and
  Systems}, pages 6106--6111, 2020.

\bibitem[Fazeli et~al.(2019)Fazeli, Oller, Wu, Wu, Tenenbaum, and
  Rodriguez]{Fazeli}
N.~Fazeli, M.~Oller, J.~Wu, Z.~Wu, J.~B. Tenenbaum, and A.~Rodriguez.
\newblock See, feel, act: Hierarchical learning for complex manipulation skills
  with multisensory fusion.
\newblock \emph{Science Robotics}, 4\penalty0 (26):\penalty0 eaav3123, 2019.

\bibitem[Ficuciello et~al.(2019)Ficuciello, Migliozzi, Laudante, Falco, and
  Siciliano]{Ficuciello}
F.~Ficuciello, A.~Migliozzi, G.~Laudante, P.~Falco, and B.~Siciliano.
\newblock Vision-based grasp learning of an anthropomorphic hand-arm system in
  a synergy-based control framework.
\newblock \emph{Science Robotics}, 4\penalty0 (26):\penalty0 eaao4900, 2019.

\bibitem[Folkestad et~al.(2020)Folkestad, Pastor, and Burdick]{Folkestad}
Carl Folkestad, Daniel Pastor, and Joel~W. Burdick.
\newblock Episodic koopman learning of nonlinear robot dynamics with
  application to fast multirotor landing.
\newblock In \emph{IEEE International Conference on Robotics and Automation},
  pages 9216--9222, 2020.

\bibitem[Ghassemi et~al.(2020)Ghassemi, Shoeibi, and Rouhani]{Ghassemi}
Navid Ghassemi, Afshin Shoeibi, and Modjtaba Rouhani.
\newblock Deep neural network with generative adversarial networks pre-training
  for brain tumor classification based on mr images.
\newblock \emph{Biomedical Signal Processing and Control}, 57:\penalty0 101678,
  2020.

\bibitem[Gilra and Gerstner(2017)]{Gilra}
Aditya Gilra and Wulfram Gerstner.
\newblock Predicting non-linear dynamics by stable local learning in a
  recurrent spiking neural network.
\newblock \emph{eLife}, 6:\penalty0 e28295, 2017.

\bibitem[Hofer et~al.(2020)Hofer, Bekris, Handa, Gamboa, Golemo, Mozifian,
  Atkeson, Fox, Goldberg, Leonard, Liu, Peters, Song, Welinder, and
  White]{hofer}
Sebastian Hofer, Kostas Bekris, Ankur Handa, Juan~Camilo Gamboa, Florian
  Golemo, Melissa Mozifian, Chris Atkeson, Dieter Fox, Ken Goldberg, John
  Leonard, C.~Karen Liu, Jan Peters, Shuran Song, Peter Welinder, and Martha
  White.
\newblock Perspectives on sim2real transfer for robotics: A summary of the rss
  2020 workshop, 2020.

\bibitem[Hu et~al.(2020)Hu, Dong, Wang, Chang, and Sun]{Hu}
Ziniu Hu, Yuxiao Dong, Kuansan Wang, Kai-Wei Chang, and Yizhou Sun.
\newblock Gpt-gnn: Generative pre-training of graph neural networks.
\newblock In \emph{ACM SIGKDD International Conference on Knowledge Discovery
  and Data Mining}, pages 1857--1867, 2020.

\bibitem[Hwangbo et~al.(2019)Hwangbo, Lee, Dosovitskiy, Bellicoso, Tsounis,
  Koltun, and Hutter]{Hwangbo}
Jemin Hwangbo, Joonho Lee, Alexey Dosovitskiy, Dario Bellicoso, Vassilios
  Tsounis, Vladlen Koltun, and Marco Hutter.
\newblock Learning agile and dynamic motor skills for legged robots.
\newblock \emph{Science Robotics}, 4\penalty0 (26):\penalty0 eaau5872, 2019.

\bibitem[Kaelbling(2020)]{Kaelbling915}
Leslie~Pack Kaelbling.
\newblock The foundation of efficient robot learning.
\newblock \emph{Science}, 369\penalty0 (6506):\penalty0 915--916, 2020.

\bibitem[Khadivar et~al.(2021)Khadivar, Lauzana, and Billard]{Khadivar}
Farshad Khadivar, Ilaria Lauzana, and Aude Billard.
\newblock Learning dynamical systems with bifurcations.
\newblock \emph{Robotics and Autonomous Systems}, 136:\penalty0 103700, 2021.

\bibitem[Levine et~al.(2018)Levine, Pastor, Alex~Krizhevsky, and
  Quillen]{Levine18}
Sergey Levine, Peter Pastor, Julian~Ibarz Alex~Krizhevsky, and Deirdre Quillen.
\newblock Learning hand-eye coordination for robotic grasping with deep
  learning and large-scale data collection.
\newblock \emph{The International Journal of Robotics Research}, 37\penalty0
  (4-5):\penalty0 421--436, 2018.

\bibitem[Liang et~al.(2020)Liang, Saxena, and Kroemer]{Liang-RSS-20}
Jacky Liang, Saumya Saxena, and Oliver Kroemer.
\newblock {Learning active task-oriented exploration policies for bridging the
  Sim-To-Real gap}.
\newblock In \emph{Robotics: Science and Systems}, 2020.

\bibitem[McKinnon and Schoellig(2017)]{Christopher}
Christopher~D. McKinnon and Angela~P. Schoellig.
\newblock Learning multimodal models for robot dynamics online with a mixture
  of gaussian process experts.
\newblock In \emph{IEEE International Conference on Robotics and Automation},
  pages 322--328, 2017.

\bibitem[Mehta et~al.(2019)Mehta, Diaz, Golemo, Pal, and Paull]{Mehta}
Bhairav Mehta, Manfred Diaz, Florian Golemo, Christopher Pal, and Liam Paull.
\newblock Active domain randomization.
\newblock In \emph{Conference on Robot Learning}, 2019.

\bibitem[Moerland et~al.(2021)Moerland, Broekens, and
  Jonker]{moerland2021modelbased}
Thomas~M. Moerland, Joost Broekens, and Catholijn~M. Jonker.
\newblock Model-based reinforcement learning: A survey.
\newblock 2021.

\bibitem[OpenAI et~al.(2019)OpenAI, Akkaya, Andrychowicz, Chociej, Litwin,
  McGrew, Petron, Paino, Plappert, Powell, Ribas, Schneider, Tezak, Tworek,
  Welinder, Weng, Yuan, Zaremba, and Zhang]{Akkaya}
OpenAI, Ilge Akkaya, Marcin Andrychowicz, Maciek Chociej, Mateusz Litwin, Bob
  McGrew, Arthur Petron, Alex Paino, Matthias Plappert, Glenn Powell, Raphael
  Ribas, Jonas Schneider, Nikolas Tezak, Jerry Tworek, Peter Welinder, Lilian
  Weng, Qiming Yuan, Wojciech Zaremba, and Lei Zhang.
\newblock Solving rubik's cube with a robot hand.
\newblock \emph{arXiv}, 2019.

\bibitem[Panda et~al.(2020)Panda, Prakash, Behera, and Dutta]{Panda}
Amrut~Sekhar Panda, Ravi Prakash, Laxmidhar Behera, and Ashish Dutta.
\newblock Combined online and offline inverse dynamics learning for a robot
  manipulator.
\newblock In \emph{International Joint Conference on Neural Networks}, pages
  1--7, 2020.

\bibitem[{Peng} et~al.(2018){Peng}, {Andrychowicz}, {Zaremba}, and
  {Abbeel}]{Peng}
Xue~Bin {Peng}, Marcin {Andrychowicz}, Wojciech {Zaremba}, and Pieter {Abbeel}.
\newblock Sim-to-real transfer of robotic control with dynamics randomization.
\newblock In \emph{IEEE International Conference on Robotics and Automation},
  pages 3803--3810, 2018.

\bibitem[Radford et~al.(2018)Radford, Narasimhan, Salimans, and
  Sutskever]{Radford}
Alec Radford, Karthik Narasimhan, Tim Salimans, and Ilya Sutskever.
\newblock Improving language understanding by generative pre-training.
\newblock \emph{Technical Report, OpenAI}, 2018.

\bibitem[Rezaei-Shoshtari et~al.(2019)Rezaei-Shoshtari, Meger, and
  Sharf]{Sahand}
Sahand Rezaei-Shoshtari, David Meger, and Inna Sharf.
\newblock Cascaded gaussian processes for data-efficient robot dynamics
  learning.
\newblock In \emph{IEEE/RSJ International Conference on Intelligent Robots and
  Systems}, pages 6871--6877, 2019.

\bibitem[Sadeghi and Levine(2017)]{Sadeghi}
Fereshteh Sadeghi and Sergey Levine.
\newblock Cad2rl: Real single-image flight without a single real image.
\newblock In \emph{Robotics: Science and Systems}, 2017.

\bibitem[Sanchez-Gonzalez et~al.(2018)Sanchez-Gonzalez, Heess, Springenberg,
  Merel, Riedmiller, Hadsell, and Battaglia]{Alvaro}
Alvaro Sanchez-Gonzalez, Nicolas Heess, Jost~Tobias Springenberg, Josh Merel,
  Martin Riedmiller, Raia Hadsell, and Peter Battaglia.
\newblock Graph networks as learnable physics engines for inference and
  control.
\newblock In \emph{International Conference on Machine Learning}, pages
  4470--4479, 2018.

\bibitem[Shah et~al.(2021)Shah, Powers, Tilton, Kriegman, Bongard, and
  Kramer-Bottiglio]{Shah}
Dylan~S. Shah, Joshua~P. Powers, Liana~G. Tilton, Sam Kriegman, Josh Bongard,
  and Rebecca Kramer-Bottiglio.
\newblock A soft robot that adapts to environments through shape change.
\newblock \emph{Nature Machine Intelligence}, 3:\penalty0 51--59, 2021.

\bibitem[Tobin et~al.(2017)Tobin, Fong, Ray, Schneider, Zaremba, and
  Abbeel]{Tobin}
Josh Tobin, Rachel Fong, Alex Ray, Jonas Schneider, Wojciech Zaremba, and
  Pieter Abbeel.
\newblock Domain randomization for transferring deep neural networks from
  simulation to the real world.
\newblock In \emph{IEEE/RSJ International Conference on Intelligent Robots and
  Systems}, pages 23--30, 2017.

\bibitem[Valassakis et~al.(2020)Valassakis, Ding, and Johns]{Valassakis}
Eugene Valassakis, Zihan Ding, and Edward Johns.
\newblock Crossing the gap: A deep dive into zero-shot sim-to-real transfer for
  dynamics.
\newblock 2020.

\bibitem[Vaswani et~al.(2017)Vaswani, Shazeer, Parmar, Uszkoreit, Jones, Gomez,
  Kaiser, and Polosukhin]{Vaswani}
Ashish Vaswani, Noam Shazeer, Niki Parmar, Jakob Uszkoreit, Llion Jones,
  Aidan~N Gomez, \L~ukasz Kaiser, and Illia Polosukhin.
\newblock {Attention is all you need}.
\newblock In \emph{Advances in Neural Information Processing Systems}, pages
  5998--6008, 2017.

\bibitem[Won et~al.(2020)Won, Müller, and Lee]{Won}
Dong-Ok Won, Klaus-Robert Müller, and Seong-Whan Lee.
\newblock An adaptive deep reinforcement learning framework enables curling
  robots with human-like performance in real-world conditions.
\newblock \emph{Science Robotics}, 5\penalty0 (46):\penalty0 eabb9764, 2020.

\bibitem[Yu et~al.(2017{\natexlab{a}})Yu, Tan, Liu, and Turk]{Wenhao}
Wenhao Yu, Jie Tan, C.~Karen Liu, and Greg Turk.
\newblock Preparing for the unknown: Learning a universal policy with online
  system identification.
\newblock In \emph{Robotics: Science and Systems}, 2017{\natexlab{a}}.

\bibitem[Yu et~al.(2017{\natexlab{b}})Yu, Tan, Liu, and Turk]{Yu-RSS-17}
Wenhao Yu, Jie Tan, C.~Karen Liu, and Greg Turk.
\newblock Preparing for the unknown: Learning a universal policy with online
  system identification.
\newblock In \emph{Robotics: Science and Systems}, 2017{\natexlab{b}}.

\bibitem[Zhou et~al.(2018)Zhou, Helwa, and Schoellig]{Zhou}
Siqi Zhou, Mohamed~K. Helwa, and Angela~P. Schoellig.
\newblock An inversion-based learning approach for improving impromptu
  trajectory tracking of robots with non-minimum phase dynamics.
\newblock \emph{IEEE Robotics and Automation Letters}, 3\penalty0 (3):\penalty0
  1663--1670, 2018.

\end{thebibliography}


\end{document}